\documentclass[10pt,twocolumn,letterpaper]{article}

\usepackage[pagenumbers]{cvpr} %

\usepackage{cuted} %

\usepackage{currfile} %

\usepackage{caption} %

\usepackage{multirow}
\usepackage{xcolor}
\usepackage{graphicx}
\usepackage{caption}
\usepackage{array}
\usepackage{adjustbox}
\usepackage{subcaption}
\usepackage[normalem]{ulem}
\usepackage[acronym]{glossaries}
\usepackage{tikz}
\usepackage{pifont}      %
\usetikzlibrary{arrows.meta,calc}

\usepackage{colortbl}
\definecolor{oursbg}{RGB}{230,239,250}     %
\definecolor{oursdark}{RGB}{200,220,246}   %
\newcommand{\pmse}[1]{{\scriptsize\color{black!45}$\pm$#1}}
\newcommand{\oursrow}{\rowcolor{oursbg}}
\newcommand{\ourscell}{\cellcolor{oursdark}}
\newcommand{\grp}{\rule{0pt}{\dimexpr\ht\strutbox+3pt\relax}}

\definecolor{oursgain}{RGB}{150,188,240}
\newcommand{\gaincap}{4}                      %
\newcommand{\gain}[1]{%
\pgfmathsetmacro\gainpct{max(0,min(100,100*(#1)/\gaincap))}%
\edef\gaincolor{\noexpand\cellcolor{oursgain!\gainpct!oursbg}}\gaincolor
\pgfmathprintnumber[fixed,fixed zerofill,precision=1,showpos]{#1}}

\usepackage{tabularx}

\usepackage{placeins}
\usepackage{bm} %

\newacronym{fixedts}{fixed TS}{fixed Temperature Scaling}
\newacronym{tta}{TTA}{Test-Time Adaptation}
\newacronym{nll}{NLL}{Negative Log-Likelihood}

\newcommand{\logit}{\operatorname{logit}}

\newcommand{\nuc}[1]{\left\lVert #1 \right\rVert_{*}}

\usepackage{xspace}

\newcommand{\Cleants}{Clean TS\xspace}

\newcommand{\Heldoutts}{Held-out TS\xspace}
\newcommand{\Fixedts}{Fixed TS\xspace}

\definecolor{cvprblue}{rgb}{0.21,0.49,0.74}
\usepackage[pagebackref,breaklinks,colorlinks,allcolors=cvprblue]{hyperref}

\title{Dynamic Duos for Reliable Test-Time Adaptation: \\Two Models are Better than One}

\author{
Alexandre St-Aubin\\
\small University of British Columbia\\
\small Vector Institute\\
{\tt\small alxstaub@cs.ubc.ca}
\and
Evan Shelhamer\\
\small University of British Columbia\\
\small Vector Institute
}

\begin{document}
\maketitle

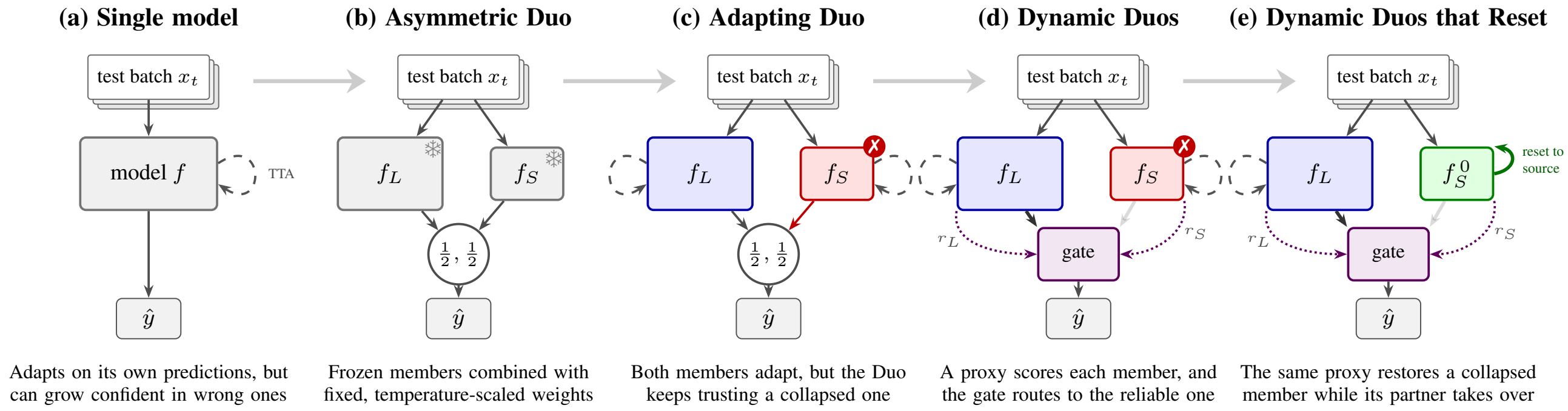
\begin{figure*}[t]
\centering
\resizebox{\textwidth}{!}{%
\begin{tikzpicture}[
    font=\footnotesize,
    >={Stealth[length=1.6mm]},
    large/.style  ={draw=blue!65!black,   fill=blue!10,   rounded corners=2pt, thick,
                    minimum width=13mm, minimum height=9mm},
    small/.style  ={draw=orange!80!black, fill=orange!15, rounded corners=2pt, thick,
                    minimum width=9mm,  minimum height=6.5mm},
    dead/.style   ={draw=red!75!black,    fill=red!12},     %
    fresh/.style  ={draw=green!50!black,  fill=green!12},   %
    largeg/.style ={draw=black!55, fill=black!6, rounded corners=2pt, thick,
                    minimum width=13mm, minimum height=9mm},
    smallg/.style ={draw=black!55, fill=black!6, rounded corners=2pt, thick,
                    minimum width=9mm,  minimum height=6.5mm},
    comb/.style   ={draw=black!70, fill=white, circle, thick, inner sep=1pt, minimum size=7.5mm},
    gate/.style   ={draw=violet!70!black, fill=violet!10, rounded corners=2pt, thick,
                    minimum width=10mm, minimum height=6.5mm},
    outbox/.style ={draw=black!70, fill=black!5, rounded corners=2pt,
                    minimum width=8mm, minimum height=5mm},
    batch/.style  ={draw=black!60, minimum width=15mm, minimum height=5.5mm, rounded corners=1pt},
    flow/.style   ={->, thick, black!70},
    bad/.style    ={->, thick, red!75!black},
    hiw/.style    ={->, black!80, line width=1.2pt},               %
    low/.style    ={->, black!80, line width=1.2pt, opacity=0.18}, %
    proxy/.style  ={->, densely dotted, violet!75!black, thick},
    tta/.style    ={->, dashed, gray!65!black, thick},
    resetarr/.style={->, green!45!black, very thick},
    ptitle/.style ={font=\small\bfseries, align=center, text width=37mm, anchor=south},
    pcap/.style   ={font=\scriptsize, align=center, text width=37mm, anchor=north,
                    execute at begin node={\hyphenpenalty=10000\exhyphenpenalty=10000}},
    note/.style   ={font=\tiny, text=black!70},
  ]

\newcommand{\batchicon}[2]{%
  \node[batch, fill=black!10] at ($#1+(0.12,-0.12)$) {};
  \node[batch, fill=black!5]  at ($#1+(0.06,-0.06)$) {};
  \node[batch, fill=white] (#2) at #1 {\scriptsize test batch $x_t$};
}
\newcommand{\collapsed}[1]{%
  \node[circle, fill=red!75!black, text=white, inner sep=0.5pt, font=\scriptsize]
    at (#1.north east) {\ding{55}};
}
\newcommand{\frozenmark}[1]{%
  \node[text=black!50, inner sep=0pt, font=\small] at ($(#1.north east)+(-0.13,-0.15)$) {\ding{100}};
}
\newcommand{\ttaloop}[2]{%
  \ifx l#2
    \draw[tta] ($(#1.west)+(0,0.13)$) .. controls +(-0.6,0.4) and +(-0.6,-0.4) .. ($(#1.west)+(0,-0.13)$);
  \else
    \draw[tta] ($(#1.east)+(0,0.13)$) .. controls +(0.6,0.4) and +(0.6,-0.4) .. ($(#1.east)+(0,-0.13)$);
  \fi
}
\newcommand{\paneltitle}[1]{%
  \node[ptitle] at (1.6,3.5) {#1};
}
\def\pw{3.85}   %

\begin{scope}[xshift=0*\pw cm]
  \paneltitle{(a) Single model}%
  \batchicon{(1.6,3.05)}{xa}
  \node[largeg, minimum width=17mm] (fa) at (1.6,1.85) {model $f$};
  \node[outbox] (ya) at (1.6,0.05) {$\hat y$};
  \draw[flow] (xa) -- (fa);
  \draw[flow] (fa) -- (ya);
  \ttaloop{fa}{r}
  \node[note, anchor=west] at ($(fa.east)+(0.5,0)$) {TTA};
  \node[pcap] at (1.6,-0.4) {Adapts on its own predictions, but can grow confident in wrong ones};
\end{scope}

\begin{scope}[xshift=1*\pw cm]
  \paneltitle{(b) \mbox{Asymmetric Duo}}%
  \batchicon{(1.6,3.05)}{xb}
  \node[largeg] (Lb) at (0.75,1.85) {$f_L$};
  \node[smallg] (Sb) at (2.45,1.85) {$f_S$};
  \frozenmark{Lb}\frozenmark{Sb}
  \node[comb]   (cb) at (1.6,0.85) {\scriptsize $\frac12,\frac12$};
  \node[outbox] (yb) at (1.6,0.05) {$\hat y$};
  \draw[flow] (xb) -- (Lb);
  \draw[flow] (xb) -- (Sb);
  \draw[flow] (Lb) -- (cb);
  \draw[flow] (Sb) -- (cb);
  \draw[flow] (cb) -- (yb);
  \node[pcap] at (1.6,-0.4) {Frozen members combined with fixed, temperature-scaled weights};
\end{scope}

\begin{scope}[xshift=2*\pw cm]
  \paneltitle{(c) Adapting Duo}%
  \batchicon{(1.6,3.05)}{xc}
  \node[large]       (Lc) at (0.75,1.85) {$f_L$};
  \node[small, dead] (Sc) at (2.45,1.85) {$f_S$};
  \collapsed{Sc}
  \ttaloop{Lc}{l}\ttaloop{Sc}{r}
  \node[comb]   (cc) at (1.6,0.85) {\scriptsize $\frac12,\frac12$};
  \node[outbox] (yc) at (1.6,0.05) {$\hat y$};
  \draw[flow] (xc) -- (Lc);
  \draw[flow] (xc) -- (Sc);
  \draw[flow] (Lc) -- (cc);
  \draw[bad]  (Sc) -- (cc);
  \draw[flow] (cc) -- (yc);
  \node[pcap] at (1.6,-0.4) {Both members adapt, but the Duo keeps trusting a collapsed one};
\end{scope}

\begin{scope}[xshift=3*\pw cm]
  \paneltitle{(d) Dynamic Duos}%
  \batchicon{(1.6,3.05)}{xd}
  \node[large]       (Ld) at (0.75,1.85) {$f_L$};
  \node[small, dead] (Sd) at (2.45,1.85) {$f_S$};
  \collapsed{Sd}
  \ttaloop{Ld}{l}\ttaloop{Sd}{r}
  \node[gate]   (gd) at (1.6,0.85) {\scriptsize gate};
  \node[outbox] (yd) at (1.6,0.05) {$\hat y$};
  \draw[flow]   (xd) -- (Ld);
  \draw[flow]   (xd) -- (Sd);
  \draw[hiw] (Ld.south) ++(0.2,0) -- (gd.north west);
  \draw[low] (Sd.south) ++(-0.15,0) -- (gd.north east);
  \draw[proxy]  (Ld.south west) to[out=-100,in=180] node[note, left, pos=0.4] {$r_L$} (gd.west);
  \draw[proxy]  (Sd.south east) to[out=-80,in=0]    node[note, right, pos=0.4] {$r_S$} (gd.east);
  \draw[flow]   (gd) -- (yd);
  \node[pcap] at (1.6,-0.4) {A proxy scores each member, and the gate routes to the reliable one};
\end{scope}

\begin{scope}[xshift=4*\pw cm]
  \node[ptitle, text width=48mm] at (1.6,3.5) {(e) Dynamic Duos that Reset};
  \batchicon{(1.6,3.05)}{xe}
  \node[large]        (Le) at (0.75,1.85) {$f_L$};
  \node[small, fresh] (Se) at (2.45,1.85) {$f_S^{\,0}$};
  \ttaloop{Le}{l}
  \node[gate]   (ge) at (1.6,0.85) {\scriptsize gate};
  \node[outbox] (ye) at (1.6,0.05) {$\hat y$};
  \draw[flow]   (xe) -- (Le);
  \draw[flow]   (xe) -- (Se);
  \draw[hiw] (Le.south) ++(0.2,0) -- (ge.north west);
  \draw[low] (Se.south) ++(-0.15,0) -- (ge.north east);
  \draw[proxy]  (Le.south west) to[out=-100,in=180] node[note, left, pos=0.4] {$r_L$} (ge.west);
  \draw[proxy]  (Se.south east) to[out=-80,in=0]    node[note, right, pos=0.4] {$r_S$} (ge.east);
  \draw[resetarr] (Se.east) to[out=10,in=-10, looseness=2.2]
     node[note, right, align=left, text=green!35!black] {reset to\\ source} (Se.north east);
  \draw[flow]   (ge) -- (ye);
  \node[pcap] at (1.6,-0.4) {The same proxy restores a collapsed member while its partner takes over};
\end{scope}

\foreach \i in {1,2,3,4} {
  \draw[-{Stealth[length=2.4mm,width=3mm]}, line width=1.4pt, black!20]
    ({\i*\pw - 0.95}, 3.0) -- ({\i*\pw + 0.45}, 3.0);
}

\end{tikzpicture}}
\caption{\textbf{From single-model adaptation to Dynamic Duos.} 
(a) Test-time adaptation updates a model on unlabeled test data, but can make it collapse: the model grows confident in wrong predictions and ends up less accurate than before adapting, with no warning.
(b) An Asymmetric Duo pairs a large model with a small sidekick and combines their scaled predictions with fixed temperatures. 
(c) When both members adapt, the fixed temperatures do not adjust for their varying accuracies. 
(d) Dynamic Duos estimate their reliability on every proxy batch with a label-free proxy, and a gate upweights the more reliable member.
(e) Dynamic Duos that Reset use the same proxy to restore a collapsed member to its source weights while the gate switches to its partner.
Gray: prior work. Dashed loops: TTA updates. Dotted arrows: proxy scores. Arrow opacity: gate weight. 
\ding{100}: frozen. \ding{55}: collapsed.
}
\label{fig:framework}
\end{figure*}

\begin{abstract}
Test-time adaptation (TTA) updates a model on unlabeled test data to improve accuracy under distribution shift. 
However, a major obstacle to deploying TTA is that it can also reduce accuracy, and most TTA methods give no signal of failure. 
We show how to adapt Asymmetric Duos, which pair a larger model with a smaller model, for more reliable adaptation with minimal overhead.
We find that a Duo is already more robust than its members even without adaptation. 
Adapting both members of the Duo can raise accuracy further, but one model can collapse while the other stays accurate, and the Duo's fixed combination keeps trusting both. 
We propose \emph{Dynamic Duos}, which replace the Duo's fixed combination with a gate that re-weights the members on every batch using a label-free reliability estimate. 
We extend gating to resetting by using the same estimate to restore a collapsed member to its source weights while its partner keeps adapting.
Dynamic Duos are compatible with any TTA update rule.
On ImageNet-C with Tent, Dynamic Duos help most where adaptation fails: they gain $5.5$ points over a fixed combination for a Duo whose small member collapses. 
Where adaptation succeeds, Independent Dynamic Duos stay within $0.2$ points of the fixed combination on every corruption.
On continual ImageNet-C, \emph{Dynamic Duos that Reset} outperform or match scheduled resets and other baselines on all settings we evaluate.
\end{abstract}

\section{Introduction}
\label{sec:intro}

Image classifiers are often trained on one distribution of images and deployed on another. 
A model trained on curated photographs can encounter images taken with other cameras, in fog or snow, or compressed for transmission, and its accuracy can drop sharply under such \emph{distribution shifts}~\cite{hendrycks2019benchmarkingneuralnetworkrobustness}. 
Labels are rarely available at deployment, so the model cannot simply be fine-tuned on the new data. 
Test-time adaptation (TTA) instead updates the model during deployment, using only the unlabeled images it receives, most often by making its predictions more confident~\cite{wang2021tentfullytesttimeadaptation,niu2022efficienttesttimemodeladaptation,niu2023towards}. 
Adaptation can recover much of the lost accuracy, but it can also leave a model less accurate than if it had not adapted at all, for instance when it grows confident in wrong answers, a failure known as \emph{collapse}~\cite{niu2022efficienttesttimemodeladaptation, zhao2023pitfallstesttimeadaptation}. 
Collapse can occur even with hyperparameters tuned on held-out shifts, and most TTA methods give no signal of it. 

Combining several models is a long-standing way to make predictions more robust, but it multiplies the cost of inference. 
Asymmetric Duos~\cite{zhou2025asymmetricduossidekicksimprove} keep much of the benefit at a fraction of the cost, pairing a large model with a much smaller \emph{sidekick} and combining their predictions with fixed temperatures.
Pairing a ViT-B/16~\cite{dosovitskiy2021image} as the large model with a ResNet-50~\cite{he2016deep} as the small model (or ``sidekick'') costs only about $25\%$ more compute than the large model alone (in forward FLOPs), yet we find it improves accuracy by 5.7 points across ImageNet-C~\cite{hendrycks2019benchmarkingneuralnetworkrobustness} corruptions (\cref{sec:exp:static}). 
Even without adaptation, a well-selected sidekick can sometimes offer a significant accuracy boost under distribution shifts.

Adapting the members of an Asymmetric Duo at test time could raise accuracy even further, but it has not been tried, and gains are not guaranteed. 
The two members differ in architecture and size, so they adapt differently, and one can collapse while the other stays accurate. 
A Duo with fixed temperatures keeps trusting the failing member, and the more confident that member becomes in its wrong predictions, the more it dominates the combination.
We propose \emph{Dynamic Duos}, which adapt both the members of an Asymmetric Duo and the way they are combined (\cref{fig:framework}). 
Throughout the stream, a lightweight gate estimates how reliable each member currently is, from its predictions on the incoming images alone, and shifts weight toward the more reliable one. 
The estimate comes from work on predicting a model's accuracy without labels~\cite{deng2023confidencedispersityspeakcharacterising}.
The gate needs no labels, no extra forward passes and no change to the underlying TTA method. 

Collapse is most damaging in continual TTA, where a model adapts over a longer stream of changing shifts rather than a single one. 
Over thousands of steps, even methods designed to resist collapse eventually degrade, and a simple baseline that outperforms many continual methods, RDumb~\cite{press2023rdumb}, resets the model to its pretrained weights at fixed intervals. 
But fixed resets are blind to the model's state. 
They throw away useful adaptation when the model is accurate and come too late when it collapses between resets. 
We introduce \emph{Dynamic Duos that Reset}, which reuse the gate's reliability estimate to decide when to reset. 
When a member's estimated reliability drops, the Duo shifts its weight to the other member and resets the failing one to its pretrained source weights.

Our contributions are as follows.
\begin{enumerate}
    \item We evaluate Asymmetric Duos under distribution shift for the first time and show that they improve robustness at little additional compute (\cref{sec:exp:static}). 
    With both models frozen, oracle temperatures fitted on test labels show how much any method that only re-weights the two models could gain (\cref{sec:exp:static}).
    \item We compare adaptation modes for Asymmetric Duos: which model adapts (the large model, the small model, or both) and which objective drives it (the Duo's joint prediction or each model's own) (\cref{sec:exp:duotta}).
    \item We propose \textit{Dynamic Duos}, which shift the Duo's combination toward whichever model remains more stable during adaptation using a reliability estimate (\cref{sub:method:dd}).
    \item We present our full method of \textit{Dynamic Duos that Reset}, a robust approach to continual TTA (\cref{sub:method:reset}) on ImageNet-C.
\end{enumerate}

\section{Background on Adaptation and Duos}
\label{sec:background}

Dynamic Duos build on two components: test-time adaptation of a single model and the Asymmetric Duo.

\paragraph{Test-time adaptation.}
Let $f_{\theta} : \mathcal{X} \to \mathbb{R}^K$ be a classifier over $K$ classes, trained on labeled data from a source distribution. 
At deployment, it receives an unlabeled stream of batches $X_1, X_2, \dots$ of $b$ samples each, drawn from a shifted target distribution, with no access to the source data or to any label. 
We write $z_t = f_{\theta_{t-1}}(X_t) \in \mathbb{R}^{b \times K}$ for the logits on batch $t$, and $P_t = \operatorname{softmax}(z_t)$ for the row-wise prediction matrix, with $p_{t,i}$ its $i$-th row. 
Adaptation is online: the model predicts on $X_t$ with parameters $\theta_{t-1}$, then updates them to $\theta_t$, without revisiting earlier samples. 
In the \emph{continual} setting, the target distribution changes over the stream, for instance from one corruption to another, at times unknown to the model.

Tent~\cite{wang2021tentfullytesttimeadaptation} minimizes the mean entropy of the predictions on each batch, $\mathcal{L}(\theta_{t-1}; X_t) = \frac{1}{b}\sum_{i} H(p_{t,i})$ with $H(p) = -\sum_{k} p_k \log p_k$, by one gradient step per batch on the affine parameters of the normalization layers, while batch normalization uses the statistics of the current test batch~\cite{schneider2020improving}.
Entropy minimization recovers much of the accuracy lost under common corruptions~\cite{hendrycks2019benchmarkingneuralnetworkrobustness} and underlies many later methods~\cite{niu2022efficienttesttimemodeladaptation,niu2023towards}.
Because it rewards confidence regardless of correctness, however, a model can reinforce its own errors until its predictions concentrate on a few classes and its accuracy falls below that of the unadapted model. 
We call this failure \emph{collapse}~\cite{niu2023towards,press2023rdumb}. 
It is most likely in the continual setting, where adaptation runs for thousands of steps across changing shifts. 

\paragraph{Asymmetric Duos.}
Ensembles of independently trained networks improve accuracy and uncertainty estimates, also under distribution shift~\cite{lakshminarayanan2017simplescalablepredictiveuncertainty,ovadia2019trustmodelsuncertaintyevaluating}, but their training and deployment costs grow linearly with the number of members. 
An Asymmetric Duo~\cite{zhou2025asymmetricduossidekicksimprove} reduces this cost to that of two models of different size: a large model $L$ and a small model $S$, the \emph{sidekick}. 
The key is to pair members with different inductive biases.
Their errors are then complementary, so even a sidekick much cheaper than the large model can improve on $L$ alone, at a fraction of the cost of a standard ensemble.
For a member $m \in \{L, S\}$, let $z^{(m)}_t$ denote its logits on batch $t$ and $P^{(m)}_t = \operatorname{softmax}(z^{(m)}_t)$ its prediction matrix. 
The Duo divides each member's logits by a temperature $T_m > 0$ and averages them,
\begin{equation}
  \bar{z}_t \;=\; \tfrac12\Big(\frac{z^{(L)}_t}{T_L} + \frac{z^{(S)}_t}{T_S}\Big),
  \label{eq:static-duo}
\end{equation}
with prediction matrix $\bar{P}_t = \operatorname{softmax}(\bar{z}_t)$. 
The temperatures are fitted once, by minimizing the negative log-likelihood of the Duo on labeled validation data, extending temperature scaling (TS) for single models~\cite{guo2017calibrationmodernneuralnetworks} to two models. 
A lower temperature sharpens a member's distribution and increases its influence on $\bar{z}_t$, so the pair $(T_L, T_S)$ both calibrates the Duo and sets the relative weight of its members, which lets the more accurate member dominate~\cite{zhou2025asymmetricduossidekicksimprove}.

\section{Dynamic Duos}
\label{sec:method}
A \emph{Dynamic Duo} adapts two things at test time: the parameters of both members (\cref{sec:duo_tent}) and the weight that combines them (\cref{sub:nuclear_norm,sub:weighting}); \emph{Dynamic Duos that Reset} (\cref{sub:method:reset}) add a label-free reset for continual streams. We start from a stronger static baseline, temperatures fitted on held-out corruptions rather than clean data (\cref{sec:motivation}).

\subsection{Recalibrating Temperatures Under Shift}
\label{sec:motivation}
Temperatures fitted on clean validation data (\emph{\Cleants})~\cite{zhou2025asymmetricduossidekicksimprove} encode how much to trust each member on clean data.
Under distribution shift, the members' logits change in ways that depend on their architecture, so their relative reliability changes too, and clean temperatures can over-weight the member that degrades more. 
We therefore fit the two temperatures to minimize the negative log-likelihood of the frozen Duo on the four held-out corruptions pooled (\emph{\Heldoutts}). 
These temperatures act as prior knowledge about how the members compare under corruption, without using any test labels.
\Cleants and \Heldoutts are both forms of \emph{fixed temperature scaling} (\emph{Fixed TS}): fitted once before the test stream and kept fixed throughout it.

To measure how much a combination that follows the shift more closely could gain, we also define oracles, which use labels and are evaluated in-sample: \emph{corruption-wise TS} refits the two temperatures on each test corruption and \emph{batch-wise TS} on every batch of $B$ samples.

\subsection{Test-time Adaptation of a Duo}
\label{sec:duo_tent}
Adapting a Duo raises two questions that single-model TTA does not: which of the members to update and which prediction to use.
Let $\mathcal{L}(z)$ denote the loss that a base TTA method minimizes on a batch of logits $z$, such as the prediction entropy for Tent~\cite{wang2021tentfullytesttimeadaptation}, or the entropy of reliable, non-redundant samples for ETA~\cite{niu2022efficienttesttimemodeladaptation}. Throughout this section, the members are combined with \Heldoutts.

First, we choose which members to update: the large one, the small one, or both. 
Adapting a single member is cheaper, since the frozen one needs only the forward pass the Duo already performs. 
Next, we choose which prediction the loss is computed on. Each adapting member's own prediction, $\mathcal{L}(z^{(m)}_t)$, as in single-model TTA, or the Duo's combined prediction under \Fixedts, $\mathcal{L}(\bar z_t)$, to which a frozen member contributes its logits as constants.
The Duo's prediction is usually more accurate than either member's~\cite{zhou2025asymmetricduossidekicksimprove}, so its loss may pull a drifting member back toward a better prediction.

Crossing the two choices gives six modes. 
With both members adapting, each can minimize its own loss, after which their predictions are combined with \Fixedts, or both can minimize the Duo's loss, which is backpropagated through both backbones. 
The single-member modes separate the effect of combining from that of the shared signal. 
In every mode, each adapting member updates the parameters its base method adapts, with that method's update, and the Duo predicts each batch from the logits computed before the update. 
But since the members differ in architecture and size, one can collapse mid-stream while the other stays accurate (\cref{sec:exp:duotta}), which fixed temperatures cannot follow.

\subsection{Measuring Member Reliability}
\label{sub:nuclear_norm}
\Heldoutts accounts for the first way the members' relative reliability changes at test time, the shift itself, before adaptation begins. 
Adaptation brings a second change that fixed temperatures cannot account for: each member's reliability evolves batch by batch, and can collapse partway through a stream. 
We address it with a gate that re-weights the two members on every proxy batch of 256 images, in two stages: a \emph{proxy} scores each member's reliability on the current batch (this section), and the gate converts the gap between the two scores into weights (\cref{sub:weighting}). 
We take the proxy from label-free accuracy estimation, which predicts a model's accuracy on unlabeled data~\cite{JMLR:v11:donmez10a}. 
Our use is finer-grained than the usual one: rather than ranking models on entire datasets, we compute the proxy on a single batch and use it only to compare two adapting members of different architectures.

Deng et al.~\cite{deng2025confidencedispersitysignalsunsupervised} group estimators that summarize softmax outputs by what they measure: the certainty of individual predictions (\emph{confidence}), the spread of predictions across classes (\emph{dispersity}), or both (\emph{hybrid}), and find hybrid estimators the most reliable. 
Our setting places two requirements on the proxy. 
First, it must detect the way TTA fails. 
Entropy minimization drives confidence up even as accuracy falls, so a collapsing member looks \emph{more} reliable to a proxy based on confidence alone, whereas its loss of dispersity, as its predictions concentrate on a few classes, reveals the collapse. 
Second, since the gate acts on the gap between the two scores, the scores must be comparable across two different architectures.

We use the \textbf{normalized nuclear norm} of the prediction matrix~\cite{deng2023confidencedispersityspeakcharacterising}, the hybrid estimator that ranked models best in~\cite{deng2025confidencedispersitysignalsunsupervised}. 
It grows with both confidence and dispersity~\cite{cui2020discriminabilitydiversitybatchnuclearnorm,deng2023confidencedispersityspeakcharacterising}, and is therefore low when predictions become either diffuse or collapsed. 
Its normalization depends only on the batch shape, so it puts both members' scores on a common scale in $[0,1]$:
\begin{equation}
  r^{(m)}_t \;=\; \frac{\nuc{P^{(m)}_t}}{\sqrt{b\,\min(b,K)}}.
  \label{nnorm}
\end{equation}
Among the ten proxies we compare (\cref{sec:exp:proxy}), only the hybrid ones meet both requirements, and the nuclear norm is the most consistent.

\subsection{Gating and Combining Predictions}
\label{sub:weighting}

The gate turns the two reliability scores of the members of the Duo into a weight $w^{(L)}_t\in(0,1)$ for the large member, with $w^{(S)}_t=1-w^{(L)}_t$:
\begin{equation}\label{eq:gate}
  w^{(L)}_t = \sigma\!\big(\beta\,\Delta_t\big),
  \qquad
  \Delta_t = \logit\big(r^{(L)}_t\big) - \logit\big(r^{(S)}_t\big),
\end{equation}
with $\beta \ge 0$. The weight depends only on the gap between the two scores, so changes that affect both members alike leave it unchanged.
It treats the members symmetrically, giving them equal weight when scores are equal and more weight to the member whose score is ahead. 
The sharpness $\beta$ sets how strongly the gate commits for a given gap. At $\beta = 0$ the weights are always $1/2$ and as $\beta$ grows it approaches hard selection of the member with the higher score. 
Because the weight varies smoothly with the gap, noise in a small gap moves it only slightly.
These weights replace the equal weights of \Fixedts (\cref{eq:static-duo}), while the temperatures stay fixed:
\begin{equation}\label{eq:combine}
  \tilde{z}_t \;=\; w^{(L)}_t\,\frac{z^{(L)}_t}{T_L} \;+\; w^{(S)}_t\,\frac{z^{(S)}_t}{T_S},
  \qquad \tilde{P}_t = \operatorname{softmax}(\tilde{z}_t).
\end{equation}
At $\beta = 0$, $\tilde{z}_t = \bar{z}_t$ and the gated Duo equals \Heldoutts. 

\subsection{Dynamic Duos}
\label{sub:method:dd}
Dynamic Duos combine \cref{sec:duo_tent,sub:weighting}: both members adapt with a base TTA method while the gate re-weights their combination on every proxy batch. 
The gate adds no forward passes and leaves the base method's loss and update unchanged, so any method can be plugged in. 
We study two variants that differ in the prediction the members adapt on.

An \emph{Independent Dynamic Duo} (\emph{Independent DD}) adapts each member on its own loss. 
The gate only changes how the members' predictions are combined, so it can be added to any pair of adapting models, and its effect can be isolated.

A \emph{Joint Dynamic Duo} (\emph{Joint DD}) adapts both members on the loss of the gated prediction, $\mathcal{L}(\tilde z_t)$ (\cref{eq:combine}). 
The weights are computed from detached logits and enter the loss as constants, so no gradient flows through the gate. 
Because the logits are pooled linearly, the gradient reaching member $m$'s logits is that of the Duo's loss scaled by $w^{(m)}_t/T_m$, so a member the gate trusts less is updated more slowly. 
This couples the gate and the adaptation, which helps when a member collapses but can also feed noise in the weights back into the members (\cref{sec:exp:dd}). 

\subsection{Dynamic Duos that Reset}
\label{sub:method:reset}

On a long stream, entropy minimization can collapse a member: its accuracy drops to a few percent and does not recover.
The gate moves weight away from a collapsed member, but that member keeps adapting on a degenerate signal and never becomes useful again.
We therefore reset a member to its source weights when its proxy signals a collapse.
The reset is triggered by the proxy the gate already computes, so it needs no labels and no extra forward pass.

Each member $m \in \{\mathrm{L},\mathrm{S}\}$ resets on its own, depending on its proxy, so neither member is ever reset because of its partner.
Dropping $m$, let $u_t = \logit(r_t)$, where $r_t$ is the member's proxy on the $t$-th proxy batch (the scale of \cref{eq:gate}).
Following the EWMA change detector~\cite{roberts1959controlcharttests,Ross_2012}, we track a fast and a slow average,
\begin{equation}
  \mu^{\mathrm{f}}_t = (1-\alpha_{\mathrm{f}})\, \mu^{\mathrm{f}}_{t-1} + \alpha_{\mathrm{f}}\, u_t, \qquad
  \mu^{\mathrm{s}}_t = (1-\alpha_{\mathrm{s}})\, \mu^{\mathrm{s}}_{t-1} + \alpha_{\mathrm{s}}\, u_t,
\end{equation}
with $0 < \alpha_{\mathrm{s}} < \alpha_{\mathrm{f}} < 1$, and flag the member when
\begin{equation}
  d_t \;=\; \frac{\mu^{\mathrm{s}}_t - \mu^{\mathrm{f}}_t}{\hat\sigma_{t-1}} \;>\; \delta ,
  \label{eq:detector}
\end{equation}
where $\hat\sigma_{t-1}$ is a running standard deviation of $u$; $\delta$ is thus measured in units of the member's own proxy noise, which puts members with different proxy scales on a common footing.
When the proxy falls, suddenly or gradually, the fast average drops before the slow one and the gap opens.
Because the test is one-sided, a rising proxy never triggers a reset.
Resets can only be triggered after a warm-up of $W_0$ proxy batches, at the start of the stream and again after every reset.
A reset restores the member's source parameters, clears its optimizer state and restarts its averages and warm-up, while the partner and the gate are left as they are.
Architectures collapse at different rates, so each member has its own learning rate $\eta_m$ and settings $\mathbf{c}_m = (\alpha_{\mathrm{f}}^{m}, \alpha_{\mathrm{s}}^{m}, W_0^{m}, \delta^{m})$.

\section{Experiments}
\label{sec:experiments}
\begin{table}
    \centering
    \footnotesize
    \setlength{\tabcolsep}{4pt}
    \setlength{\aboverulesep}{0pt}
    \setlength{\belowrulesep}{0pt}
    \setlength{\extrarowheight}{1.5pt}
    \caption{\textbf{Temperatures under shift.} Top-1 accuracy (\%) of the two frozen Duos on source/clean validation data and averaged over target/test corruptions. Held-out TS closes much of the gap between clean TS and the corruption-wise oracle on frozen Duos. Bold: best without labels. Shaded: oracles that use test labels. Optimal temperatures in \cref{fig:temp_ratios}, per-corr. results in \cref{tab:temperatures:frozen_bn__vitb16_rn50v1,tab:temperatures:frozen_bn__vitb16_convnextt}.
    }
    \label{tab:temperatures:duos}
    \begin{tabularx}{\linewidth}{@{}Xcccc@{}}
        \toprule
        & \multicolumn{2}{c}{ViT-B/16 + RN50} & \multicolumn{2}{c}{ViT-B/16 + CNX-T} \\
        \cmidrule(lr){2-3}\cmidrule(l){4-5}
                                    & Source & Target & Source & Target \\
        \midrule
        Large alone                 & 81.1 & 40.4 & 81.1 & 40.4 \\
        Small alone                 & 75.2 & 31.6 & 82.5 & 29.8 \\
        \midrule
        Fixed TS on clean data      & \textbf{81.5} & 45.6 & \textbf{83.4} & 38.1 \\
        Fixed TS on held-out corr.  & \textbf{81.5} & \textbf{46.1} & 83.0 & 41.0 \\
        \midrule
        \rowcolor{gray!12} Oracle TS per corruption  & 81.5 & 46.8 & 83.4 & 42.2 \\
        \rowcolor{gray!12} Oracle TS per batch of 16 & 82.4 & 47.4 & 84.3 & 42.7 \\
        \bottomrule
    \end{tabularx}
\end{table}

\label{sec:exp:setup}

\paragraph{Models and data.}
Our main Duo pairs a ViT-B/16~\cite{dosovitskiy2021image} (large, LayerNorm) with a ResNet-50~\cite{he2016deep} (small, BatchNorm). 
A second Duo replaces the ResNet-50 with a ConvNeXt-T~\cite{liu2022convnet2020s}, a small model of similar size from a different family. 
All models use the torchvision \texttt{IMAGENET1K\_V1} weights. We evaluate on the 15 ImageNet-C corruptions~\cite{hendrycks2019benchmarkingneuralnetworkrobustness} at severity~5 (50{,}000 images each) under two protocols. 
In the \emph{episodic} protocol, both members are reset to their source weights at every corruption boundary. 
In the \emph{continual} protocol~\cite{wang2022continual}, the corruptions follow one another without resets, in the standard order and four of CoTTA's random orders. The ResNet-50 always normalizes with the statistics of the current test batch (test-time BN~\cite{schneider2020improving}), so even our frozen ResNet-50 is mildly adapted.
With its source statistics it averages only $18\%$ at severity~5, and every TTA method we study starts from this model, so temperatures fitted on it remain valid once adaptation starts.
The ViT-B/16 and ConvNeXt-T use LayerNorm and are frozen entirely.

\paragraph{Adaptation.}
Each member adapts with Tent~\cite{wang2021tentfullytesttimeadaptation}, ETA~\cite{niu2022efficienttesttimemodeladaptation}, or SAR~\cite{niu2023towards}, using SGD (momentum $0.9$) with one gradient step per batch of 64. 
The gate uses $\beta = 0.25$ unless stated otherwise, and we also report other values for Independent DD in \cref{app:beta}. 
The proxy is computed on batches of $256$ samples, independently of the adaptation batch size, which balances its reliability against the delay in detecting a collapse.

\paragraph{Baselines.}
\emph{\Fixedts} is the fixed combination of \cref{eq:static-duo} with held-out temperatures (\cref{sec:motivation}), \ie the gated Duo at $\beta = 0$.
On continual streams, we compare when each member is reset: \emph{Never}; RDumb~\cite{press2023rdumb}, every 1{,}000 steps; ASR~\cite{lim2026resetmatterslongtermtesttime} and SAR's model recovery, at their published thresholds (for ASR, those for continual ImageNet-C, also used for the ViT-B/16); and \emph{Random}, which resets each member exactly as many times as our rule, at uniformly random steps, and so differs from ours only in timing.
Combining the two members' logits, with the gate or with \Fixedts, gives a Duo version of each baseline at no extra cost.

Following Rusak et al.~\cite{rusak2023datadistributionshiftsuse}, we tune all hyperparameters, and fit \Heldoutts, on the four held-out ImageNet-C corruptions~\cite{hendrycks2019benchmarkingneuralnetworkrobustness} (speckle noise, spatter, Gaussian blur and saturate), which appear in no test stream (details in \cref{app:tuning}).

\subsection{Static Duos Under Shift}
\label{sec:exp:static}
\label{sec:exp:temps}

\begin{figure}
    \centering
    \includegraphics[width=0.95\linewidth]{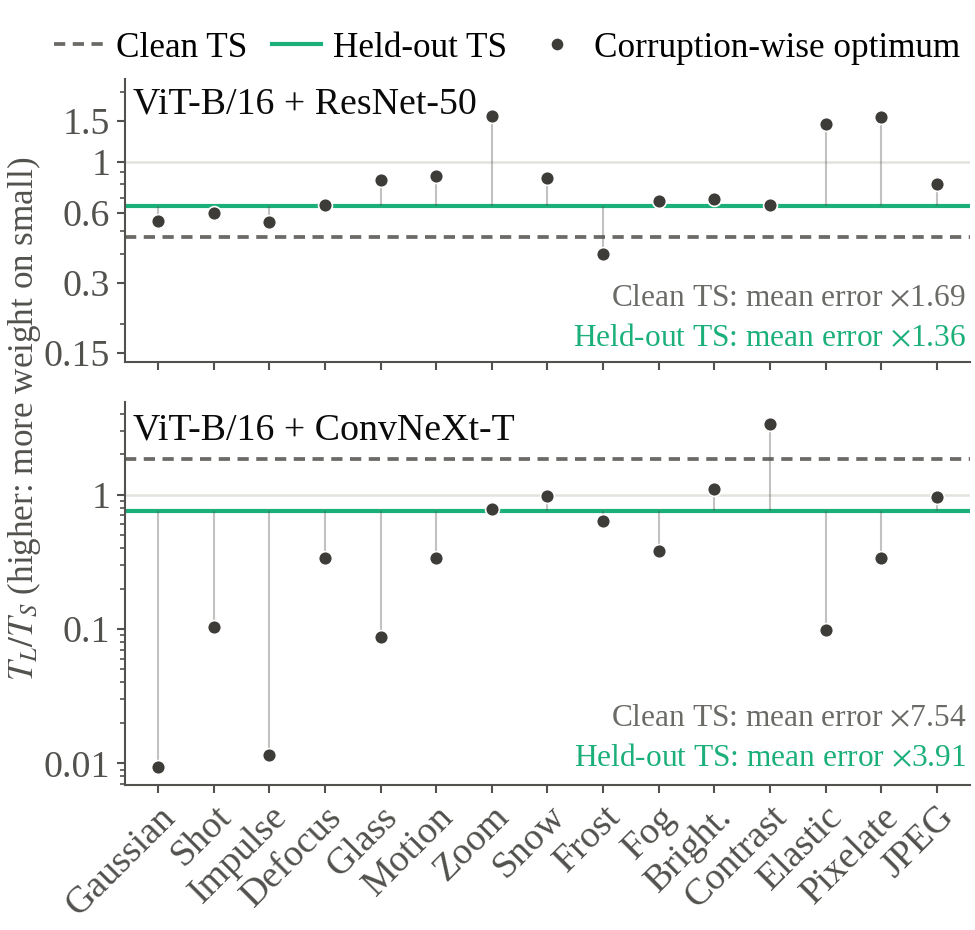}
    \caption{\textbf{The optimal temperature ratio moves across corruptions.} NLL-optimal ratio $T_L^*/T_S^*$ of each frozen Duo on each test corruption (log scale), which defines the corruption-wise oracle of \cref{tab:temperatures:duos}.
    We report each fixed ratio's geometric mean error: the factor by which it misses the optimum, averaged over test corruptions, where $\times 1$ would be exact.
    Held-out TS moves the ratio from its clean value toward the optima on both Duos.}
    \label{fig:temp_ratios}
\end{figure}

\Cref{tab:temperatures:duos} compares frozen Duos with their members and with oracle temperatures fitted on test labels (\cref{sec:motivation}).

\paragraph{A Duo is more robust than its members if weighted correctly.}
With Clean TS, the ResNet-50 Duo already beats both members on all 15 test corruptions, and Held-out TS raises its lead over the ViT to $5.7$ points ($46.1\%$ vs.\ $40.4\%$) at no cost on clean data.
The ConvNeXt-T Duo shows why the weighting matters.
ConvNeXt-T is the stronger member on clean data ($82.5\%$ vs.\ $81.1\%$), so Clean TS gives it more weight ($T_L/T_S = 1.85$), but it degrades more under shift ($29.8\%$ vs.\ $40.4\%$) and drags the Duo below the ViT alone ($38.1\%$).
Held-out TS reverses the weighting ($T_L/T_S = 0.77$) and lifts the Duo above the ViT ($41.0\%$).
On the ResNet-50 Duo, it raises the ratio from $0.47$ to $0.64$, gaining most on pixelate, zoom blur and elastic transform, where the ResNet-50 overtakes the ViT.

\paragraph{Held-out TS approaches the oracle.}
Held-out TS moves the temperature ratio from its clean value toward the corruption-wise optima on both Duos, reducing its mean multiplicative error from $\times 1.69$ to $\times 1.36$ on the ResNet-50 Duo and from $\times 7.54$ to $\times 3.91$ on the ConvNeXt-T Duo (\cref{fig:temp_ratios}). 
This closes much of the accuracy gap between clean TS and the corruption-wise oracle: from $45.6\%$ to $46.1\%$ (oracle: $46.8\%$) on the ResNet-50 Duo, and from $38.1\%$ to $41.0\%$ (oracle: $42.2\%$) on the ConvNeXt-T Duo. 

Within a corruption, by contrast, there is little left to follow. Refitting the temperatures on every batch of 16 images gains at most $0.6$ points over the corruption-wise oracle, and even more on clean data ($0.9$), where there is no shift to track, so this gain comes from fitting sampling noise. On frozen Duos, a fixed combination fitted on held-out corruptions is therefore close to the best achievable with one temperature pair per corruption.

\subsection{Adapting Duos}
\label{sec:exp:adapt_vs_duo}
\label{sec:exp:duotta}
\label{sec:exp:dd}

\begin{table}
    \centering
    \small
    \setlength{\tabcolsep}{3pt}
    \caption{\textbf{Adapting, making a Duo, or both.} Top-1 accuracy (\%) on ImageNet-C at severity 5, averaged over the 15 corruptions, with batch size 64. 
    Mean over 3 seeds. 
    \emph{Adapted}: Tent, each model on its own entropy. 
    \emph{Duo}: ViT-B/16 with a ResNet-50 (RN50) or ConvNeXt-T (CNX-T), combining their temperature-scaled logits with Fixed TS. 
    Making a Duo helps every individual model, adapting helps all but ConvNeXt-T, and doing both is best.
    }
    \label{tab:adapt_or_duo}
    \begin{tabular*}{\linewidth}{@{\extracolsep{\fill}}lccccc@{}}
        \toprule
        & \multicolumn{3}{c}{Single model} & \multicolumn{2}{c}{Duo: ViT-B/16 +} \\
        \cmidrule(lr){2-4}\cmidrule(l){5-6}
                & ViT-B/16 & RN50 & CNX-T & RN50 & CNX-T \\
        \midrule
        Frozen  & 40.4 & 31.7 & 29.8 & 46.1 & 41.0 \\
        Adapted & 42.9 & 43.9 & 27.3 & 52.2 & 43.2 \\
        \bottomrule
    \end{tabular*}
\end{table}

\begin{table}
    \centering
    \small
    \setlength{\tabcolsep}{3pt}
    \caption{\textbf{How to adapt a Duo.} Top-1 accuracy (\%) on ImageNet-C at severity 5, averaged over the 15 corruptions, with Tent and batch size 64. The gate uses $\beta{=}0.25$. 
    Mean over 3 seeds. 
    Best result per Duo in bold. 
    We investigate which model(s) to adapt and which entropy to optimize.
    Adapting both models is best across both Duos, and which entropy depends on the Duo.
    \emph{Own}/\emph{Duo}: each adapted model minimizes the entropy of its own prediction or of the Duo's prediction.
    }
    \label{tab:adaptation_modes_avg:tent__vitb16_rn50v1_vs_vitb16_convnextt_gate}
    \begin{tabular*}{\linewidth}{@{\extracolsep{\fill}}llcccc@{}}
        \toprule
        & & \multicolumn{2}{c}{ViT-B/16 + RN50} & \multicolumn{2}{c}{ViT-B/16 + CNX-T} \\
        \cmidrule(lr){3-4}\cmidrule(l){5-6}
        Model(s) & Combination           & Own  & Duo  & Own  & Duo  \\
        \midrule
        Large    & Fixed TS              & 48.6 & 56.2 & 44.2 & 39.4 \\
        Small    & Fixed TS              & 50.6 & 50.5 & 37.9 & 36.5 \\[2pt]
        Both     & Fixed TS              & 52.2 & \textbf{59.1} & 43.2 & 38.6 \\
        Both     & Gate & 52.5 & 58.1 & \textbf{45.8} & 44.1 \\
        \bottomrule
    \end{tabular*}
\end{table}

\begin{figure*}[t]
  \centering
  \includegraphics[width=0.9\textwidth]{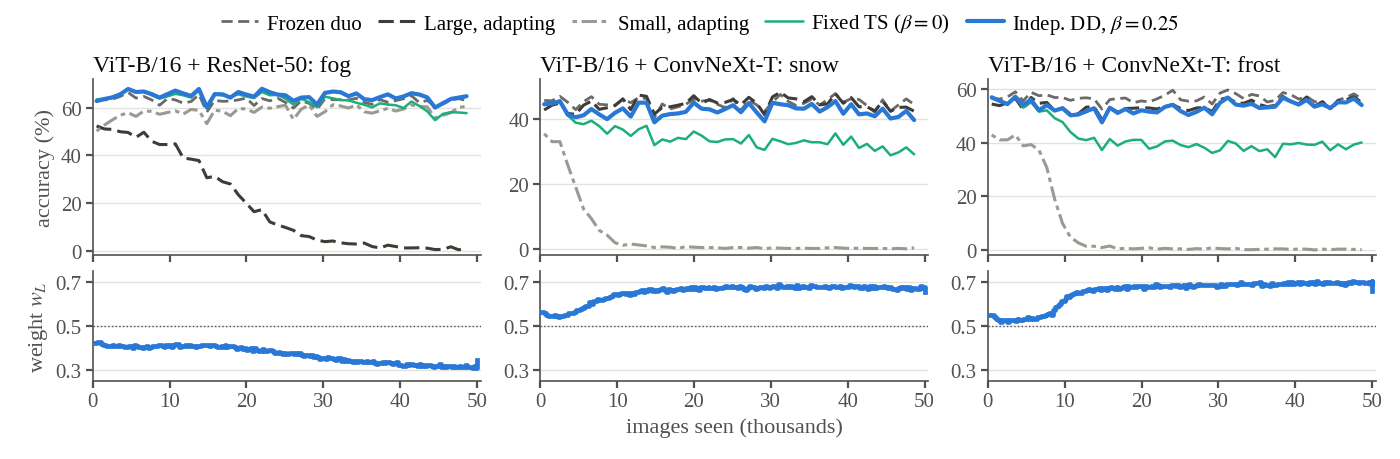}
  \caption{\textbf{Gate behaviour when one Duo member collapses.}
Single ImageNet-C shifts (severity 5, 50k images, episodic) from the Independent Dynamic Duo (DD) runs of \cref{tab:adaptation_modes_avg:tent__vitb16_rn50v1_vs_vitb16_convnextt_gate}, with our gate at $\beta = 0.25$ and Fixed TS ($\beta = 0$).
\emph{Top:} accuracy. \emph{Bottom:} gate weight on the large member.
The gate recovers much of the accuracy of the stable member when the other collapses, while Fixed TS does not.
}
  \label{fig:gate-stress}
\end{figure*}

We now adapt the members of the Duo with Tent, under the modes of \cref{sec:duo_tent}. \Cref{tab:adaptation_modes_avg:tent__vitb16_rn50v1_vs_vitb16_convnextt_gate} combines the members either with \Fixedts (\Heldoutts) or with our gate. 
Modes that adapt with our gate reuse the learning rates tuned with \Fixedts, so any difference between the two comes from the gate. 
Single-member modes that use their own entropy reuse the rates tuned for individual models. All other modes are tuned on the Duo's accuracy (\cref{tab:hparams_episodic}).
\Cref{tab:adapt_or_duo} compares with adapting a single model; per-corruption results are in \cref{tab:adaptation_modes:tent__vitb16_rn50v1_bn_gate_nnn_bt256_beta0p25-0p5_recpair,tab:adaptation_modes:tent__vitb16_convnextt_bn__worst_shift_gate_nnn_bt256_beta0p25-0p5_recpair}.

\paragraph{A stable Duo.}
Adding a ResNet-50 sidekick to the ViT-B/16 and adapting both help, and they add up: the frozen Duo ($46.1\%$) beats either model adapted alone ($42.9\%$ and $43.9\%$), and adapting both members inside it, each on its own entropy, reaches $52.2\%$.
Adapting both members on the Duo entropy is best by a wide margin ($59.1\%$).
Single-model Tent collapses the ViT on fog ($19.5\%$ vs.\ $51.2\%$ frozen). 
However, next to a frozen ResNet-50, adapting the ViT on its own entropy gives a Duo at $55.8\%$ on fog, and adapting it on the Duo entropy $67.6\%$.
Outside the adaptation loop, the gate is safe: Independent DD is never worse than Fixed TS on any corruption and gains $0.3$ points ($52.5\%$), most on fog ($+2.0$), where the ViT collapses (\cref{fig:gate-stress}, left).
Inside the loop, in Joint DD, it costs $1.0$ point ($58.1\%$). 

\paragraph{An unstable Duo.}
Even at rates tuned on held-out corruptions, ConvNeXt-T collapses under Tent on snow, frost, fog and elastic transform ($3.3\%$ to $9.2\%$), and ends below its frozen accuracy ($27.3\%$ vs.\ $29.8\%$).
A fixed combination keeps trusting it.
Adapting both members on their own entropy ($43.2\%$) does worse than adapting the ViT alone inside the Duo ($44.2\%$), and adapting both on the Duo entropy falls below the frozen Duo ($38.6\%$ vs.\ $41.0\%$), collapsing on contrast, zoom blur and snow ($2.4\%$ to $11.5\%$).
The gate turns this around, adding $2.6$ points in Independent DD ($45.8\%$) and $5.5$ in Joint DD ($44.1\%$).
The gains come from the corruptions where a member collapses, up to $+11.5$ on frost in Independent DD and $+27.6$ on snow in Joint DD, and the gate never loses more than $0.4$ points on any corruption.
As ConvNeXt-T collapses, the gate moves its weight to the ViT and tracks its accuracy, while Fixed TS settles well below the survivor (\cref{fig:gate-stress}, middle and right).
Adapting both members now pays off: Independent DD is the best result on this Duo, $2.9$ points above the best adapted single model.

\paragraph{The gate routes, it does not repair.}
The best mode depends on the Duo, but in both, failures come from a member that degrades while the combination keeps trusting it.
The gate fixes this only while one member still works.
When both collapse, as with the ConvNeXt-T on fog in Independent DD, it cannot recover the Duo.
Repairing a collapsed member requires resetting it (\cref{sec:exp:reset}).

\begin{table}
  \centering\footnotesize
  \setlength{\tabcolsep}{3.5pt}
  \begin{tabular}{@{}>{\columncolor{white}[0pt][\tabcolsep]}llcc>{\columncolor{oursbg}}c>{\columncolor{oursbg}[\tabcolsep][0pt]}c@{}}
    \toprule
    Method & Reset & ViT-B/16 & ResNet-50 & \multicolumn{1}{c}{DD (ours)} & \multicolumn{1}{c}{$\Delta$} \\
    \midrule
    \multicolumn{2}{@{}l}{No adaptation} & 40.4 & 31.7 & 45.9 & $+$5.5 \\[3pt]
    Tent & None         & 45.1\pmse{0.3} & 38.4\pmse{0.3} & 50.4\pmse{0.2} & $+$5.3 \\
         & Random       & 44.5\pmse{1.3} & 39.6\pmse{0.3} & 51.6\pmse{1.0} & $+$7.1 \\
         & RDumb        & 43.8\pmse{0.4} & \underline{42.0}\pmse{0.2} & 51.7\pmse{0.1} & $+$7.9 \\
         & SAR recovery & 48.8\pmse{0.8} & 34.2\pmse{1.7} & 52.9\pmse{0.3} & $+$4.1 \\
         & ASR          & 50.6\pmse{0.3} & 38.4\pmse{0.2} & 54.4\pmse{0.2} & $+$3.8 \\
    \oursrow & DD (ours)     & \underline{50.7}\pmse{0.5} & 38.8\pmse{0.4} & \ourscell\textbf{54.5}\pmse{0.1} & \ourscell$+$3.8 \\
    \grp ETA & None     & 52.4\pmse{0.2} & 44.2\pmse{0.1} & 55.5\pmse{0.1} & $+$3.1 \\
             & Random       & 55.8\pmse{0.1} & 46.3\pmse{0.1} & 58.5\pmse{0.1} & $+$2.7 \\
             & RDumb        & 56.5\pmse{0.1} & 46.6\pmse{0.2} & 58.9\pmse{0.1} & $+$2.4 \\
             & ASR          & 37.1\pmse{8.5} & \underline{47.4}\pmse{0.1} & 55.8\pmse{1.9} & $+$8.4 \\
    \oursrow & DD (ours)     & \underline{57.5}\pmse{0.0} & 47.3\pmse{0.2} & \ourscell\textbf{59.8}\pmse{0.1} & \ourscell$+$2.3 \\
    \grp SAR & Random       & 48.3\pmse{2.9} & 35.1\pmse{3.3} & 51.9\pmse{1.8} & $+$3.6 \\
         & RDumb        & 51.5\pmse{0.9} & \underline{44.7}\pmse{0.3} & \textbf{56.5}\pmse{0.3} & $+$5.0 \\
         & SAR recovery & 42.8\pmse{9.7} & 40.5\pmse{0.3} & 52.5\pmse{3.0} & $+$9.7 \\
    \oursrow & DD (ours)     & \underline{53.4}\pmse{0.2} & 40.7\pmse{0.6} & \ourscell 56.2\pmse{0.2} & \ourscell$+$2.8 \\
    \bottomrule
  \end{tabular}
  \caption{\textbf{Single-model TTA vs.\ the Duo with and without resetting.} 
  ViT-B/16 + ResNet-50 with Independent DD with $\beta = 0.25$ on continual ImageNet-C: top-1 accuracy (\%), averaged over the 15 corruptions at severity 5, mean $\pm$ standard error over 5 corruption orders. 
  $\Delta$: Duo minus its better member. 
  Bold: best Duo per method. 
  Underline: best single model per method. 
  For DD reset rule, see~\cref{sub:method:reset}.
  Across the update and reset methods, the Duo always improves over the single best model. 
  The Duo without adaptation uses our gate ($45.9\%$), $0.2$ points below Fixed TS (\cref{tab:temperatures:duos}).
  }
  \label{tab:duo_vs_single}
\end{table}

\subsection{Dynamic Duos that reset: continual adaptation}
\label{sec:exp:reset}
Section~\ref{sec:exp:duotta} reset both members at every known corruption boundary. 
We now remove that knowledge and evaluate Dynamic Duos that reset (Section~\ref{sub:method:reset}) on the continual protocol, which is what the reset rule is for. 
Within a single corruption, the gate can down-weight a collapsing member, but over many more samples, that member keeps degrading and only a reset may bring it back.

\paragraph{The Duo helps every TTA method.}
We compare Independent DD with its two members running alone: each member adapts and resets on its own, and the gate combines their logits, requiring no additional tuning over what is needed for the single models. 
Under every TTA method and every reset scheme, the Duo beats its better member, by $2.3$ to $9.7$ points (\cref{tab:duo_vs_single}). 
The gain is largest when one member is unstable. 
With ETA and ASR's reset, the ViT-B/16 falls to $37.1\%$, with a standard error of $8.5$ that reflects collapses in some corruption orders, yet the Duo still reaches $55.8\%$. Two models are better than one.

\paragraph{When to reset.}
Our resetting scheme gives the best Duo under Tent and ETA, and under SAR it beats SAR's own recovery by $3.7$ points and trails RDumb by $0.3$, within one standard error. 
It improves on never resetting, the worst choice under both Tent and ETA, by $4.1$ and $4.3$ points, whereas the other rules each do well under only one method. 
ASR is level with ours under Tent ($54.4\%$ vs.\ $54.5\%$) but gains only $0.3$ points over never resetting under ETA, where its spread across orders is large, and RDumb comes second under ETA but adds just $1.3$ points under Tent. 

The individual members respond differently to these rules. 
Ours gives the best ViT-B/16 under all three TTA methods, whereas for the ResNet-50, RDumb is better under Tent and SAR, and ASR is level with ours under ETA. 
We attribute this to how each model fails. 
The ViT-B/16 collapses abruptly, so its proxy drops sharply and the comparison of a fast and a slow moving average catches it. 
The ResNet-50 drifts slowly, its proxy barely moves, and the rule fires late, while a fixed schedule resets it regardless. 

The gains do not come from resetting the right number of times: \emph{Random}, which differs from ours only in timing, trails it by $2.9$ points with Tent and $1.3$ with ETA, so \textit{when} a member resets matters, not only \textit{how often}. The same conclusions hold for a second Duo (\cref{tab:second_duo}).

\paragraph{How the members should adapt.}
When both members adapt on the gated prediction, what one learns changes what the other is trained on (\cref{tab:indep_vs_joint}). 
Without resets, this coupling leaves the mean unchanged ($50.5\%$ vs.\ $50.4\%$) but makes the Duo much less stable across orders, with a standard error of $2.4$ instead of $0.2$. 
With resets, joint adaptation helps: it gains $2.4$ points over independent adaptation with our trigger, $5.1$ with RDumb. 
Timing also matters more. 
Resetting as often as ours but at random steps drops the joint Duo to $46.5\%$, close to no adaptation ($45.9\%$) and $10.4$ points below our trigger. 
Our trigger, however, no longer beats a fixed schedule: it is level with RDumb ($56.9\%$ vs.\ $56.8\%$). 
When both members adapt on one shared signal, they may degrade together, and we suspect each member's proxy then declines too gradually for the fast/slow comparison to fire early.
Joint adaptation is thus the more accurate setup when it is well tuned, but also the harder one to tune. 
Its members share one loss, so their learning rates must be searched together, as a pair, for each reset rule, and the search interacts with the gate, whose weights change what each member learns. 
Independent DD needs none of this: each member keeps the learning rate and reset setting tuned for it alone, as in single-model TTA, and the gate, with one $\beta$ fixed for all experiments, is added on top without any retuning. 
\begin{table}[t]
  \centering\footnotesize
  \setlength{\tabcolsep}{3pt}
  \begin{tabular}{@{}lcc@{\hspace{6pt}}cc@{}}
    \toprule
    & \multicolumn{2}{c}{Single model} & \multicolumn{2}{c}{Duo} \\
    \cmidrule(lr){2-3}\cmidrule(l){4-5}
    Reset & ViT-B/16 & ResNet-50 & Independent & Joint \\
    \midrule
    No adaptation & 40.4 & 31.7 & \multicolumn{2}{c}{45.9} \\[2pt]
    None   & 45.1\pmse{0.3} & 38.4\pmse{0.3} & 50.4\pmse{0.2} & 50.5\pmse{2.4} \\
    RDumb  & 43.8\pmse{0.4} & 42.0\pmse{0.2} & 51.7\pmse{0.1} & \textbf{56.8}\pmse{0.1} \\
    Random & 44.5\pmse{1.3} & 39.6\pmse{0.3} & \textbf{51.6}\pmse{1.0} & 46.5\pmse{2.4} \\[2pt]
    \oursrow DD (ours) & 50.7\pmse{0.5} & 38.8\pmse{0.4} & 54.5\pmse{0.1} & \textbf{56.9}\pmse{0.2} \\
    \bottomrule
  \end{tabular}
  \caption{\textbf{Independent vs.\ joint adaptation of the Duo.} ViT-B/16 + ResNet-50 with Tent on continual ImageNet-C: top-1 accuracy (\%) averaged over the 15 corruptions at severity 5, mean $\pm$ standard error over 5 corruption orders. \emph{Single model}: each member adapting alone on its own entropy, with the same reset rule, taken from the Independent runs. 
  Bold: the better Duo setup per row.
  Joint runs with our trigger and Random use $\beta = 0$, and the other Joint runs use $\beta = 0.25$ (\cref{app:tuning}).
  }
  \label{tab:indep_vs_joint}
\end{table}

\section{Related Work}
\label{sec:related}

\paragraph{Test-time adaptation of ensembles.}
\label{sec:related:tta_ensembles}
Few works adapt more than one model at test time. Mean teacher~\cite{tarvainen2018meanteachersbetterrole} and self-ensembling~\cite{french2018selfensemblingvisualdomainadaptation} keep a moving average of a network as a pseudo-labeling teacher, as do several continual TTA methods~\cite{wang2022continual,dobler2023robustmeanteachercontinual}, and BACS~\cite{zhou2021trainingtestdatabayesian} and TTA-DAME~\cite{jeon2026ttadametesttimeadaptationdomain} adapt ensemble members without coordination; in all of them, the members share one architecture.
The closest work is COCA~\cite{yi2025smallguideslargecrossmodel}, which adapts a large anchor and a small auxiliary model, including our ViT-B/16 and ResNet-50, on the entropy of a combination that rescales the auxiliary's logits to agree with the anchor's.
Dynamic Duos differ in three ways. COCA's weighting is one-sided and presumes the anchor is reliable, whereas our gate scores each member independently and can move weight away from the large model, which is the member that collapses on snow and frost in COCA's own baselines~\cite[Table~1]{yi2025smallguideslargecrossmodel}. Our gate also leaves the base method's objective unchanged, and our reset handles continual streams without known shift boundaries.

\paragraph{Continual TTA and resetting.}
\label{sec:related:reset}
On continual streams~\cite{wang2022continual}, a model's errors can accumulate into collapse (\cref{sec:background}). Most continual methods pull the model toward its source parameters~\cite{wang2022continual,niu2022efficienttesttimemodeladaptation,marsden2023universaltesttimeadaptationweight,hoang2024persistenttesttimeadaptationrecurring}, yet on long streams most fall below the unadapted model and are beaten by periodic resets~\cite{press2023rdumb}.
Resetting methods differ in when they reset: on a fixed schedule (RDumb~\cite{press2023rdumb}), when the running entropy falls below a threshold (SAR~\cite{niu2023towards}), on a severe shift from the source (DA-TTA~\cite{wang2024distributionalignmentfullytesttime}), or when predictions concentrate on a few classes, resetting only the last layers (ASR~\cite{lim2026resetmatterslongtermtesttime}).
Each discards adaptation and must predict with the restored parameters while it re-adapts, whereas a Duo can reset one member while the gate relies on the other (\cref{sub:method:reset}).

\paragraph{Calibration under shift.}
The calibration of temperature scaling~\cite{guo2017calibrationmodernneuralnetworks} deteriorates under shift~\cite{ovadia2019trustmodelsuncertaintyevaluating}. Remedies fit the temperature on synthetically shifted data~\cite{tomani2021posthocuncertaintycalibrationdomain}, as our held-out TS does, on several training domains~\cite{yu2022robustcalibrationmultidomaintemperature}, or on a calibration set reweighted toward the target~\cite{park2020calibratedpredictioncovariateshift}. In a Duo, temperatures also weight the members, and under TTA the right weighting changes within a stream.

\section{Conclusion}
\label{sec:conclusion}
Test-time adaptation can silently leave a model worse than before it adapted.
We showed that Asymmetric Duos make adaptation more reliable and proposed \emph{Dynamic Duos}, which re-weight their members throughout the stream with a label-free reliability estimate, and \emph{Dynamic Duos that Reset}, which use it to restore a collapsed member.
Neither needs labels, extra forward passes or a modified TTA method.

Dynamic Duos beat each member adapted alone in every setting we test.
With Tent in the episodic setting, Independent DD exceeds the best single model by $8.6$ points with ViT-B/16 + ResNet-50 and $2.9$ with ViT-B/16 + ConvNeXt-T, whose small member collapses.
In the continual setting, the main Duo beats its better member by $2.3$ to $9.7$ points under every TTA method and reset rule.
Independent DD is safe and easy to use: its members keep the settings tuned for them alone, the gate adds one $\beta$ fixed across experiments, and in the episodic setting it never costs more than $0.2$ points on any corruption while gaining up to $11.5$ where a member collapses.
Tuning $\beta$ on held-out shifts further limits this risk, as $\beta = 0$ recovers \Fixedts.
Our reset rule is the only one within one standard error of the best under all three TTA methods, and improves on never resetting by $4.1$ points with Tent and $4.3$ with ETA.
Adapting both members on the gated prediction can be more accurate but needs jointly tuned learning rates.

Several limitations remain. 
The best $\beta$ varies across shifts and Duos, so adapting it during the stream is a natural next step. 
We also tested only entropy minimization TTA methods that adapt on each batch, while there are more types of optimization for TTA.
Our reset rule detects a slowly drifting member like the ResNet-50 late, and no longer beats a fixed schedule when both members adapt on the gated prediction and degrade together. 
Finally, our continual experiments cover only ImageNet-C, leaving longer streams and natural shifts to future work.

\section*{Acknowledgments}
We would like to thank Geoff Pleiss and Tim G. Zhou for their valuable advice throughout the project and thank Tim for the feedback on the exposition.

AST completed much of this work while supported by the NSRC CGRS-M.
ES is supported by a Canada CIFAR AI Chair and the Natural Sciences and Engineering Research Council of Canada (NSERC) Discovery Grant (RGPIN-2025-06878).
Resources used in preparing this research were provided, in part, by the Province of Ontario, the Government of Canada through CIFAR, and \href{https://vectorinstitute.ai/partnerships/current-partners/}{companies sponsoring} the Vector Institute.

\section*{LLM Usage Statement}
We used Claude (Anthropic), a large language model, in the writing of this paper. It helped draft and condense text, suggested rewordings, and checked that the numbers quoted in the text matched our tables. It also helped with some tables and the TikZ code of the framework figure. It also assisted with code, by writing parts of our experiment and table-generation scripts, which we reviewed and tested. All experiments were designed and run by the authors, and every reported number comes from our own runs. We reviewed all text, code and figures produced with the model's help, and verified every citation, and we take full responsibility for the content of the paper.

{
    \small
    \bibliographystyle{ieeenat_fullname}
    \bibliography{ref/main, ref/baselines}
}
\clearpage
\appendix
\makeatletter
\setlength{\@fptop}{0pt}
\setlength{\@fpsep}{\floatsep}
\setlength{\@fpbot}{0pt plus 1fil}
\makeatother

\newcommand{\tablepair}[3][1]{%
  \begin{table*}[p]
    \vbox to \textheight{%
      \vfill
      \begingroup
      \renewenvironment{table*}[1][]{}{\par}%
      \let\origresizebox\resizebox
      \def\resizebox##1##2{\origresizebox{#1##1}{##2}}%
      \input{#2}%
      \vspace{2\floatsep}%
      \input{#3}%
      \endgroup
      \vfill}%
  \end{table*}}

\maketitlesupplementary

\section{Hyperparameter Tuning}
\label{app:tuning}
\begin{table}[b]
  \centering\footnotesize
  \setlength{\tabcolsep}{4pt}
  \caption{\textbf{Settings shared by all experiments.} Temperatures are $(T_L, T_S)$, fitted on the frozen Duo. \Heldoutts is used everywhere except the clean-TS rows of \cref{tab:temperatures:duos}.}
  \label{tab:hparams_shared}
  \begin{tabularx}{\linewidth}{@{}l>{\raggedright\arraybackslash}X@{}}
    \toprule
    Data & ImageNet-C, severity 5, $50{,}000$ images per corruption, native $224{\times}224$ input \\
    Weights & torchvision \texttt{IMAGENET1K\_V1}; test-batch BN for the ResNet-50 \\
    Optimizer & SGD, momentum $0.9$, one step per batch of 64 \\
    Gate & nuclear norm on batches of 256, $\beta = 0.25$ \\
    \midrule
    \multicolumn{2}{@{}l}{\emph{Temperatures} $(T_L, T_S)$ (\cref{sec:motivation})} \\
    \quad ResNet-50 Duo & clean $(0.69, 1.46)$, held-out $(0.65, 1.01)$ \\
    \quad ConvNeXt-T Duo & clean $(1.11, 0.60)$, held-out $(0.72, 0.94)$ \\
    \midrule
    ETA & entropy margin $E_0 = 0.4 \ln 1000$, redundancy margin $\epsilon = 0.05$, decay $0.9$ \\
    SAR & $\rho = 0.05$, entropy margin $E_0 = 0.4 \ln 1000$ \\
    SAR recovery & published ($e_0 = 0.2$, decay $0.9$) \\
    ASR & published continual ImageNet-C thresholds for RN50 and ViT-B/16 \\
    \bottomrule
  \end{tabularx}
\end{table}

All hyperparameters are tuned on the four held-out ImageNet-C corruptions (speckle noise, spatter, Gaussian blur and saturate), never on a test stream, except for one choice described at the end of this section.
In the episodic protocol, each candidate is scored by its online accuracy on the four held-out streams.
Settings that work on held-out shifts can still fail on others~\cite{zhao2023pitfallstesttimeadaptation}. 
ConvNeXt-T is more prone to collapse, so for the ViT-B/16 + ConvNeXt-T Duo we use a more conservative rule: among the candidates within $1$ point of the best mean accuracy, we keep the one with the highest worst-stream accuracy. 
In the continual protocol, each candidate is scored by its mean accuracy on a stream that visits the four held-out corruptions four times in rotated order (16 visits, 800{,}000 images).
This approximates the length of the test stream at 15 corruptions of 50,000 images.
Single models are tuned on their own accuracy. 
In a Duo, an adapting member is tuned on the Duo's accuracy, with the other member frozen, or jointly when both adapt.
In continual Independent DD, each member's trajectory depends only on its own rate and reset times, so each member is tuned on its own accuracy.
\begin{table}[b]
  \centering\footnotesize
  \setlength{\tabcolsep}{4pt}
  \caption{\textbf{Episodic learning rates} (Tent), large / small member; -- marks a frozen member. Joint DD reuses the rates of both members adapting on the Duo entropy. For the ConvNeXt-T Duo, rates are selected with the worst-stream criterion of \cref{app:tuning}.}
  \label{tab:hparams_episodic}
  \begin{tabular}{@{}lcc@{}}
    \toprule
    Adapting, entropy & ResNet-50 Duo & ConvNeXt-T Duo \\
    \midrule
    Single model, own        & 7e-5 / 5e-4     & 7e-5 / 2.2e-6 \\[2pt]
    Large, own               & 7e-5 / --       & 7e-5 / -- \\
    Small, own               & -- / 5e-4       & -- / 2.2e-6 \\
    Both, own (Indep.\ DD)   & 7e-5 / 5e-4     & 1e-4 / 3.1e-6 \\[2pt]
    Large, Duo               & 1.12e-3 / --    & 5.6e-4 / -- \\
    Small, Duo               & -- / 1.4e-3     & -- / 5e-5 \\
    Both, Duo (Joint DD)     & 3.2e-3 / 4e-3   & 6.2e-6 / 5e-5 \\
    \bottomrule
  \end{tabular}
\end{table}

\paragraph{Resetting.}
The running standard deviation $\hat\sigma$ of \cref{eq:detector} is an exponentially weighted standard deviation of the member's score with rate $0.01$, started from the variance of the $W_0^m$ warm-up batches. 
It restarts with the averages. 
Our reset rule (\cref{sub:method:reset}) has four settings per member: the smoothing factors $\alpha_f^m, \; \alpha_s^m$ of its fast and slow averages, the length of its warm-up $W_0^m$, and its threshold $\delta^m$. 
We tune them, together with the member's learning rate, on the held-out streams only, in two stages.
First, we replay the rule offline on proxy sequences recorded from held-out runs in which the members collapse and runs in which they stay healthy, and keep a few settings that detect collapse quickly while rarely firing on healthy runs. 
Second, we run each candidate on the held-out stream, retuning the learning rate for it, and among the candidates within one standard error of the best accuracy, we keep the one that resets least. 
In Joint DD, the learning rates, the gate's $\beta$ and the reset settings of our rule are tuned in turn on the Duo's accuracy. 
Held-out tuning picks $\beta = 0$, so our rule and its Random control run at $\beta = 0$ in \cref{tab:indep_vs_joint}. 
Every other Joint DD configuration runs at the fixed $\beta = 0.25$ (\cref{app:beta}).
For ETA and SAR, our rule uses the learning rates tuned for RDumb, so that only the timing of the resets differs. 
For Tent with Independent DD, the reported rule in \Cref{tab:duo_vs_single} uses thresholds lower than the held-out-tuned ones. 
The held-out-tuned rule reaches $54.1 \pm 0.0$, against $54.5 \pm 0.1$ for the reported threshold. 
With it, our rule is within $0.3$ points of the best reset rule under Tent (ASR, $54.4 \pm 0.2$) and under SAR (RDumb), and best under ETA, so it is the only rule within $0.3$ points of the best under all three methods.
The lower thresholds help on the test stream because they catch the slow drift of the ResNet-50 earlier, a drift that hardly occurs on the held-out stream at its learning rate. 

The baseline reset rules keep their published settings (\cref{tab:hparams_shared}), with the learning rates of \cref{tab:hparams_continual}. 
With ASR's published ViT-B/16 configuration (from its CCC experiments), ETA + ASR reaches $40.7\%$ on the ViT-B/16: it resets about every four batches and stays close to the frozen model ($40.4\%$).
We therefore use ASR's continual ImageNet-C thresholds, with the ViT-B/16's learning rate tuned on the held-out stream. 
SAR is evaluated with its own model recovery, with RDumb's schedule and with our rule. 
We did not run it without resets or with ASR's reset, whose detector and thresholds are defined for ETA.
The exact hyperparameters used in our experiments are specified in \Cref{tab:hparams_continual,tab:hparams_episodic,tab:hparams_shared}.

\section{Choice of the Gate Sharpness \texorpdfstring{$\beta$}{beta}}
\label{app:beta}
The proxy is computed on windows of $256$ test images, four adaptation batches, from the logits the Duo already computes for its predictions. 
The weights are recomputed each time a window completes, and the batch that completes it is combined with the new weights. 
The other batches use the weights of the last completed window. 
Before the first window completes, $w^{(L)} = w^{(S)} = 1/2$, which in the episodic protocol happens at the start of every corruption.

The sharpness trades the gate's reactiveness against staying close to the fixed combination.
A larger $\beta$ moves more weight for a given gap, so the gate turns away from a collapsing member sooner and more decisively.
When both members stay stable, however, the gap is mostly noise and the fixed combination is the better choice, so a larger $\beta$ then costs accuracy.
Which side of this trade-off matters most depends on how often adaptation makes a member fail, and that varies from one shift to the next, since learning rates that are safe on most corruptions can still collapse a member on a few (\cref{sec:exp:duotta}), and the held-out corruptions need not contain the failures that occur at test time.
On the held-out corruptions, the ResNet-50 Duo is insensitive to $\beta$ up to $0.5$, while the ConvNeXt-T Duo, whose small member is unstable, prefers $\beta$ between $0.5$ and $1$ (\cref{tab:indep_dd_beta_sweep}).
We fixed $\beta = 0.25$ once, before any continual test run, as the largest $\beta$ that was never measurably below Fixed TS on the held-out continual stream of the ResNet-50 Duo ($+0.02 \pm 0.02$).
It was not retuned per Duo, stream or configuration.
For Joint DD, $\beta$ also affects adaptation: at $\beta = 0.5$, the ResNet-50 Duo reaches $55.5\%$ instead of $58.1\%$ and the ConvNeXt-T Duo $45.2\%$ instead of $44.1\%$ (\cref{tab:adaptation_modes:tent__vitb16_rn50v1_bn_gate_nnn_bt256_beta0p25-0p5_recpair,tab:adaptation_modes:tent__vitb16_convnextt_bn__worst_shift_gate_nnn_bt256_beta0p25-0p5_recpair}).

\begin{table}
    \centering
    \small
    \setlength{\tabcolsep}{5pt}
    \begin{tabular}{lcccc}
        \toprule
        & \multicolumn{4}{c}{ViT-B/16 +} \\
        \cmidrule(lr){2-5}
        & \multicolumn{2}{c}{ResNet-50} & \multicolumn{2}{c}{ConvNeXt-T} \\
        \cmidrule(lr){2-3}\cmidrule(lr){4-5}
        $\bm{\beta}$ & Held-out & Test & Held-out & Test \\
        \midrule
        0 (Fixed TS) & \textbf{56.6} & 52.2 & 55.3 & 43.2 \\
        0.125 & \textbf{56.6} & 52.4 & 55.6 & 45.3 \\
        0.25 & \textbf{56.6} & \textbf{52.5} & 55.9 & 45.8 \\
        0.5 & \textbf{56.6} & \textbf{52.5} & \textbf{56.1} & \textbf{46.0} \\
        1 & 56.5 & \textbf{52.5} & \textbf{56.1} & 45.9 \\
        2 & 56.4 & \textbf{52.5} & 55.6 & 45.5 \\
        4 & 56.1 & 52.2 & 54.8 & 45.0 \\
        \bottomrule
    \end{tabular}
        \caption{Independent DD at several levels of the gate sharpness $\beta$: top-1 accuracy (\%) averaged over the corruptions (at severity 5, in the episodic setting, with Tent, and the mean over 3 seeds).
        The best per column is in bold across the models and evaluation sets, with \emph{Held-out} for the 4 tuning corruptions and \emph{Test} for the 15 test corruptions.
        The best sharpness on held-out is indicative of the best sharpness on test, and accuracy is fairly insensitive to the exact value: for $\beta$ between $0.125$ and $2$, it varies by at most $0.2$ points on the ResNet-50 Duo and $0.7$ points on the ConvNeXt-T Duo.
        The gate of \cref{eq:gate} (proxy batches of 256) is applied outside the adaptation loop, so every $\beta$ is replayed on the same member logits.
        $\beta = 0$ is the Fixed TS Duo.
        }
    \label{tab:indep_dd_beta_sweep}

\end{table}

\section{A Second Duo}
We evaluate the gains of a second Duo (ViT-B/16 and ConvNeXt-T) on continual ImageNet-C in \Cref{tab:second_duo}.
\begin{table}[t]
  \centering\footnotesize
  \setlength{\tabcolsep}{3.5pt}
  \begin{tabular}{@{}>{\columncolor{white}[0pt][\tabcolsep]}lcc>{\columncolor{oursbg}}c>{\columncolor{oursbg}[\tabcolsep][0pt]}c@{}}
    \toprule
    Reset strategy         & ViT-B/16 & ConvNeXt-T & \multicolumn{1}{c}{Duo (ours)} & \multicolumn{1}{c}{$\Delta$} \\
    \midrule
    No adaptation          & 40.4 & 29.8 & 41.5 & $+$1.1 \\
    None                  & 45.7\pmse{0.3} & 4.5\pmse{1.1} & 46.1\pmse{0.1} & $+$0.4 \\
    Random                 & 43.9\pmse{1.2} & 23.8\pmse{0.4} & 45.2\pmse{1.2} & $+$1.3 \\
    RDumb    & 44.7\pmse{0.5} & 24.6\pmse{0.9} & 46.1\pmse{0.6} & $+$1.4 \\[2pt]
    \oursrow DD (ours) & \textbf{50.0}\pmse{0.5} & \textbf{29.1}\pmse{0.2} & \ourscell\textbf{51.4}\pmse{0.5} & \ourscell$+$1.4 \\
    \bottomrule
  \end{tabular}
\caption{\textbf{A second Duo: ViT-B/16 + ConvNeXt-T.} 
    Tent with Independent DD on continual ImageNet-C, same protocol as \cref{tab:duo_vs_single}, except that hyperparameters are selected by worst-shift accuracy (the best worst held-out stream among candidates within 1 point of the best mean) instead of mean accuracy for more robustness. 
    This tuning is necessary because ConvNeXt-T is more prone to collapse.
    ViT-B/16 differs from \cref{tab:duo_vs_single} only because of this selection rule. 
    The Duo with no adaptation uses our gate and reaches $41.5\%$, $0.5$ points over the Fixed TS combination. 
    Bold: best continual row per column. 
    }
  \label{tab:second_duo}
\end{table}

\section{Proxies}
\label{app:proxies} 
For member $m$ and proxy batch $t$ of $b_t$ images, $P^{(m)}_t\in[0,1]^{b_t\times K}$ is the matrix whose $i$-th row is $p_i=\mathrm{softmax}(z_i)$. Here $z_i$ are the member's raw logits (no temperature) and $K=|\mathcal{C}|$ is the number of classes. We write
$H(p)=-\sum_{c\in\mathcal{C}}p_c\log p_c$ for the entropy,
$\bar p=\frac{1}{b_t}\sum_{i}p_i$ for the batch-mean prediction,
$\|\cdot\|_*$ and $\|\cdot\|_F$ for the nuclear and Frobenius norms,
and $n_t=\min(b_t,K)$.
Every proxy is transformed so that a higher value means a more reliable member, and is clipped to $[0,1]$. 

\subsection{Confidence}
These proxies measure how certain each row of the prediction matrix $P$ is. We expect these to perform well under well-behaved adaptation streams, but to be confidently wrong during a collapse episode. 
\paragraph{Average confidence (ConfScore).}
The simplest confidence signal is the maximum softmax probability, which Hendrycks and Gimpel~\cite{hendrycks2018baselinedetectingmisclassifiedoutofdistribution} use to detect misclassified and out-of-distribution examples. Averaged over a set of predictions, it becomes the baseline accuracy estimator of~\cite{guillory2021predictingconfidenceunseendistributions,deng2025confidencedispersitysignalsunsupervised}:
\begin{equation}
r^{(m)}_t=\frac{1}{b_t}\sum_{i=1}^{b_t}\max_{c}\;\big(P^{(m)}_t\big)_{ic}.
\end{equation}

\paragraph{Negative entropy.}
One minus the average prediction entropy, normalized to $[0,1]$~\cite{guillory2021predictingconfidenceunseendistributions}:
\begin{equation}
r^{(m)}_t=1-\frac{1}{b_t\log K}\sum_{i=1}^{b_t}H(p_i).
\end{equation}
Tent minimizes this entropy~\cite{wang2021tentfullytesttimeadaptation}, so under Tent the score rises with adaptation whether or not accuracy does.

\paragraph{Average Thresholded Confidence (ATC)~\cite{garg2022leveragingunlabeleddatapredict}.}
A per-member threshold $\tau^{(m)}$ is fitted once, on the frozen member, using a labeled slice of clean ImageNet training images (10 images per class, $10{,}000$ images): $\tau^{(m)}$ is the quantile of the slice's scores at level $1-\mathrm{acc}^{(m)}_{\mathrm{src}}$, where $\mathrm{acc}^{(m)}_{\mathrm{src}}$ is the member's top-1 accuracy on the slice, so that the fraction of slice samples whose score reaches $\tau^{(m)}$ equals the member's source accuracy. 
The threshold then stays fixed while the member adapts, and the proxy is the fraction of the batch that reaches it:

\begin{equation}
r^{(m)}_t=\frac{1}{b_t}\sum_{i=1}^{b_t}\mathbb{I}\big[s(p_i)\geq\tau^{(m)}\big].
\end{equation}
We report both scores proposed by Garg et al.~\cite{garg2022leveragingunlabeleddatapredict}, each with its own threshold: the maximum confidence $s(p)=\max_c p_c$, ATC (conf.), and the negative entropy $s(p)=-H(p)$, ATC (ent.).

\paragraph{AvgEnergy.}
The negative mean free energy of the batch \cite{liu2021energybasedoutofdistributiondetection, deng2025confidencedispersitysignalsunsupervised}:
\begin{equation}
r^{(m)}_t=\frac{1}{b_t}\sum_{i=1}^{b_t} T\log\sum_{c\in\mathcal C}\exp\!\big(z^{(m)}_{ic}/T\big),\qquad T=1.
\end{equation}
Deng et al.~\cite{deng2025confidencedispersitysignalsunsupervised} define AvgEnergy with the opposite sign; we negate it so that, like every other proxy, a higher score means higher predicted accuracy. 

\begin{table*}[t]
\centering
\footnotesize
\setlength{\tabcolsep}{4pt}
\caption{How well each label-free proxy tracks accuracy, on the held-out continual streams of Independent DD (both members adapt with Tent, held-out shifts only), grouped by family as in \cref{app:proxies}. \emph{Healthy}: both members at stable learning rates. 
\emph{Collapsing}: at rates high enough that both collapse. 
Spearman correlation over the proxy batches of one stream, averaged over the 4 streams per Duo and regime. 
\emph{Gap}: the gate's input, the difference of the two members' scores, against the difference of their accuracies. 
$L$, $S$: each member's score against its own accuracy. 
An asterisk denotes the proxy we use. 
Best per column in bold.}
\label{tab:proxy_correlation}
\begin{tabular}{l ccc ccc@{\hspace{5\tabcolsep}}ccc ccc}
\toprule
 & \multicolumn{6}{c}{ViT-B/16 (L) + ResNet-50 (S)} & \multicolumn{6}{c}{ViT-B/16 (L) + ConvNeXt-T (S)} \\
\cmidrule(lr){2-7} \cmidrule(lr){8-13}
 & \multicolumn{3}{c}{Healthy} & \multicolumn{3}{c}{Collapsing} & \multicolumn{3}{c}{Healthy} & \multicolumn{3}{c}{Collapsing} \\
\cmidrule(lr){2-4} \cmidrule(lr){5-7} \cmidrule(lr){8-10} \cmidrule(lr){11-13}
Proxy & Gap & $L$ & $S$ & Gap & $L$ & $S$ & Gap & $L$ & $S$ & Gap & $L$ & $S$ \\
\midrule
\multicolumn{13}{l}{\textit{Confidence}} \\
\quad ConfScore~\cite{guillory2021predictingconfidenceunseendistributions,deng2025confidencedispersitysignalsunsupervised} & .62 & .89 & .94 & $-$.68 & $-$.68 & $-$.52 & .63 & .88 & .84 & $-$.45 & $-$.07 & $-$.32 \\
\quad Neg.\ entropy~\cite{guillory2021predictingconfidenceunseendistributions} & .64 & .83 & .92 & $-$.72 & $-$.70 & $-$.52 & .59 & .82 & .82 & $-$.45 & $-$.11 & $-$.33 \\
\quad ATC (conf.)~\cite{garg2022leveragingunlabeleddatapredict} & .55 & .90 & .94 & $-$.52 & $-$.68 & $-$.50 & .64 & \textbf{.90} & .84 & $-$.42 & $-$.06 & $-$.38 \\
\quad ATC (ent.)~\cite{garg2022leveragingunlabeleddatapredict} & .53 & .87 & .94 & $-$.34 & $-$.67 & $-$.49 & .47 & .87 & .83 & $-$.39 & $-$.08 & $-$.38 \\
\quad AvgEnergy~\cite{liu2021energybasedoutofdistributiondetection,deng2025confidencedispersitysignalsunsupervised} & .57 & .59 & .87 & $-$.25 & $-$.74 & $-$.55 & .23 & .58 & .72 & $-$.50 & $-$.26 & $-$.35 \\
\midrule
\multicolumn{13}{l}{\textit{Diversity}} \\
\quad Marg.\ entropy~\cite{deng2025confidencedispersitysignalsunsupervised} & $-$.28 & $-$.68 & $-$.92 & .89 & .72 & .71 & .54 & $-$.68 & .07 & .71 & .20 & .57 \\
\quad ASR conc.~\cite{lim2026resetmatterslongtermtesttime} & $-$.45 & .85 & .20 & .69 & .80 & .83 & .76 & .85 & .82 & .88 & .88 & .78 \\
\midrule
\multicolumn{13}{l}{\textit{Hybrid}} \\
\quad Nuclear norm$^\ast$~\cite{cui2020discriminabilitydiversitybatchnuclearnorm,deng2023confidencedispersityspeakcharacterising} & \textbf{.67} & .90 & .94 & \textbf{.97} & \textbf{.82} & \textbf{.95} & .85 & .89 & \textbf{.96} & \textbf{.94} & \textbf{.92} & \textbf{.82} \\
\quad Mutual info.~\cite{liang2020reallyneedaccesssource} & .65 & .84 & .92 & .96 & .80 & .93 & \textbf{.86} & .84 & .94 & \textbf{.94} & .87 & .81 \\
\midrule
\multicolumn{13}{l}{\textit{Agreement}} \\
\quad Duo agreement~\cite{chen2023detectingerrorsestimatingaccuracy} & .64 & \textbf{.91} & \textbf{.97} & $-$.49 & $-$.14 & .36 & .82 & .88 & \textbf{.96} & .20 & .01 & .51 \\
\bottomrule
\end{tabular}
\end{table*}

\subsection{Diversity}

\paragraph{Marginal entropy.}
The normalized entropy of the batch-mean prediction, the ClassEntropy of~\cite{deng2025confidencedispersitysignalsunsupervised}. It is high when the batch's predictions are spread over many classes and falls when a member collapses onto a few:
\begin{equation}
r^{(m)}_t=\frac{H(\bar p)}{\log K}.
\end{equation}

\paragraph{ASR prediction concentration.}
ASR~\cite{lim2026resetmatterslongtermtesttime} triggers resets on the \emph{prediction concentration} $C_t=\sum_{c}\hat p_{t,c}\log\hat p_{t,c}$, the negative entropy of $\hat p_t=\mathrm{softmax}\big(\frac{1}{b_t}\sum_{i=1}^{b_t}z_i\big)$, the softmax of the batch-mean \emph{logits}, where a large $C_t$ signals concentrated predictions and a risk of collapse. We use it with the opposite orientation, normalized to $[0,1]$, so that higher means more reliable:
\begin{equation}
r^{(m)}_t=-\frac{C_t}{\log K}=\frac{1}{\log K}\,H\!\Big(\mathrm{softmax}\Big(\frac{1}{b_t}\sum_{i=1}^{b_t}z_i\Big)\Big).
\end{equation}

\subsection{Hybrid}
\paragraph{Mutual information.}
The mutual information between the input and the predicted class, estimated on the batch. It is the information-maximization objective~\cite{liang2020reallyneedaccesssource}: diversity, the entropy of the mean prediction, minus the mean entropy of the individual predictions, so it also combines confidence and diversity:
\begin{equation}
r^{(m)}_t=\frac{1}{\log n_t}\Big(H(\bar p)-\frac{1}{b_t}\sum_{i=1}^{b_t}H(p_i)\Big),
\end{equation}
set to $0$ when $n_t=1$.

\paragraph{Normalized nuclear norm~\cite{deng2023confidencedispersityspeakcharacterising, cui2020discriminabilitydiversitybatchnuclearnorm}.}
The normalized nuclear norm of the prediction matrix captures both confidence and prediction diversity \cite{deng2025confidencedispersitysignalsunsupervised}. It is large only when the predictions are both confident and spread across many classes:
\begin{equation}
r^{(m)}_t=\frac{\big\|P^{(m)}_t\big\|_*}{\sqrt{b_t\,n_t}}.
\end{equation}

\subsection{Agreement}
\paragraph{Duo agreement.}
The fraction of the batch on which the member's predicted class matches that of the fixed (ungated) Duo combination $z^{\mathrm{Duo}}_i=\tfrac12 z^{(L)}_i/T_L+\tfrac12 z^{(S)}_i/T_S$, adapted from ensemble-agreement accuracy estimators~\cite{chen2023detectingerrorsestimatingaccuracy}:
\begin{equation}
r^{(m)}_t=\frac{1}{b_t}\sum_{i=1}^{b_t}\mathbb{I}\Big[\arg\max_c z^{(m)}_{ic}=\arg\max_c z^{\mathrm{Duo}}_{ic}\Big].
\end{equation}
In Chen et al.~\cite{chen2023detectingerrorsestimatingaccuracy}, the reference ensemble is made of separately trained models. Here, the member contributes half of the reference prediction. With two members, $z^{\mathrm{Duo}}$ usually predicts the class of one of them, so the gap in agreement mostly reflects which member dominates the combined logits.

\begin{figure*}
    \caption{\textbf{Accuracy vs.\ compute.} 
    Episodic top-1 accuracy (\%) on ImageNet-C at severity 5, averaged over the 15 corruptions, against GFLOPs per image for one adaptation step with Tent (forward, backward and gate). 
    Circles: single models. 
    Squares: Duos with frozen members (gray) or Dynamic Duos. 
    Triangles: one member adapting on the Duo entropy. 
    On the ResNet-50 Duo, Joint DD beats Large Tent by 15.2 points for only $1.24\times$ the FLOPs.
    }
    \label{fig:acc_vs_flops}

    \centering
    \begin{subfigure}{0.49\textwidth}
        \centering
        \includegraphics[width=\linewidth]{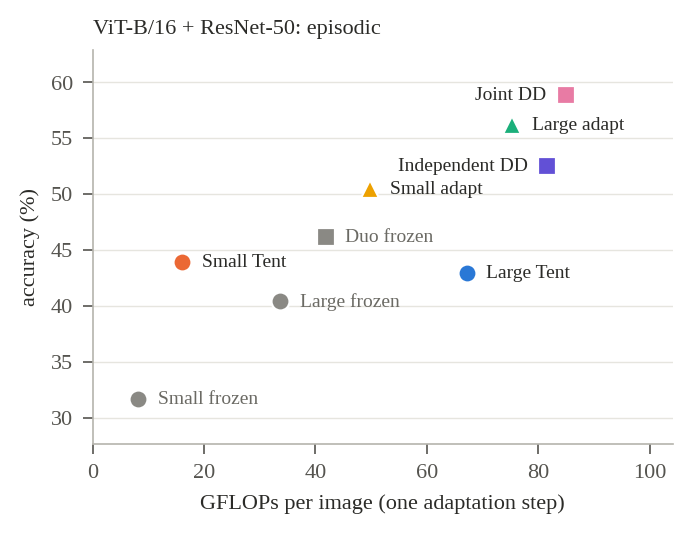}
        \caption{ViT-B/16 + ResNet-50.}
        \label{fig:acc_vs_flops_vitb_rn50}
    \end{subfigure}\hfill
    \begin{subfigure}{0.49\textwidth}
        \centering
        \includegraphics[width=\linewidth]{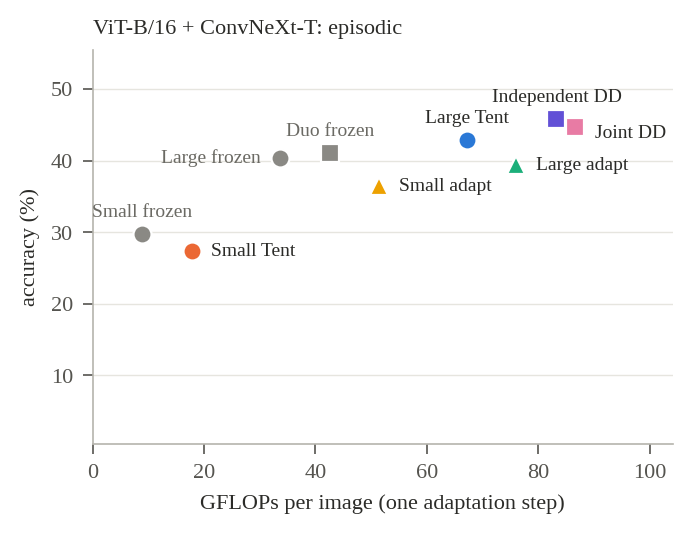}
        \caption{ViT-B/16 + ConvNeXt-T.}
        \label{fig:acc_vs_flops_vitb_convnext}
    \end{subfigure}
\end{figure*}

\subsection{Choice of proxy}
\label{sec:exp:proxy}

We compare the ten proxies defined above against the two requirements of \cref{sub:nuclear_norm}: a proxy must keep tracking a member's accuracy when that member collapses, and the gap between the two members' scores must track the gap in their accuracies, although the members are different architectures.

We run Independent DD with Tent on continual streams built only from the four held-out corruptions, in four orderings, and in two regimes per Duo. On \emph{healthy} streams, both members adapt at the learning rates tuned in \cref{sec:exp:duotta}, at which neither collapses. On \emph{collapsing} streams, the rates are raised until both members collapse. Because the members never see the gate in Independent DD, every proxy is evaluated on the same trajectories, on every proxy batch of 256 images. For each stream, we compute two Spearman correlations over proxy batches: between each member's score and its own accuracy, and between the gap in scores and the gap in accuracies, which is what the gate acts on. Correlations are computed within each stream and then averaged over streams, for the two Duos of the previous sections.

Confidence alone does not work as a reliability signal under adaptation (\cref{tab:proxy_correlation}). On healthy streams, ConfScore, negative entropy and ATC track each member's accuracy well (Spearman $.82$ to $.94$), with gap correlations of $.47$ to $.64$. 
On collapsing streams, their gap correlation turns negative on both Duos, between $-.72$ and $-.34$. 
A collapsing member grows \emph{more} confident as it degrades, so a gate driven by confidence would shift weight toward the member that is failing. 
Diversity proxies avoid this failure, since collapse removes diversity, but they compare two architectures poorly: on the healthy ViT-B/16 + ResNet-50 streams, the gap correlations of marginal entropy and ASR's concentration are negative ($-.28$ and $-.45$), although ASR's concentration tracks the ViT-B/16's own accuracy well ($.85$). Tracking each member separately is therefore not enough. 
The two hybrid proxies are the only ones whose gap correlation stays above $.6$ for both Duos in both regimes. 
The nuclear norm is the best, or within $.01$ of the best for the gap: $.67$ and $.97$ on healthy and collapsing streams for ViT-B/16 + ResNet-50, and $.85$ and $.94$ for ViT-B/16 + ConvNeXt-T. Mutual information is close behind everywhere, and slightly ahead on the healthy ConvNeXt-T streams. We use the nuclear norm, the most consistent across Duos and regimes.

\section{Computation cost}
\begin{table}[t]
  \centering\small
  \setlength{\tabcolsep}{4pt}
  \begin{tabular}{@{}l cc @{\hspace{12pt}} cc@{}}
    \toprule
    & \multicolumn{2}{c}{+ ResNet-50} & \multicolumn{2}{c}{+ ConvNeXt-T} \\
    \cmidrule(lr){2-3}\cmidrule(lr){4-5}
    Method & Acc.(\%) & GFLOPs & Acc.(\%) & GFLOPs \\
    \midrule
    Large frozen          & 40.4 & 33.7 & 40.4 & 33.7 \\
    Small frozen          & 31.7 &  8.2 & 29.8 &  8.9 \\
    \addlinespace
    Large Tent             & 42.9 & 67.2 & 42.9 & 67.2 \\
    Small Tent             & 43.9 & 16.1 & 27.3 & 17.8 \\
    \addlinespace
    Duo frozen~\cite{zhou2025asymmetricduossidekicksimprove} & 46.1 & 41.9 & 41.0 & 42.6 \\
    Indep. DD (ours)       & 52.5 & 83.3 & \textbf{45.8} & 85.0 \\
    \textbf{Joint DD} (ours)      & \textbf{58.1} & 83.3 & 44.1 & 85.0 \\
    \midrule
    \emph{Ours vs.\ Large Tent} & $+$15.2 & $+$24\% & $+$2.9 & $+$26\% \\
    \bottomrule
  \end{tabular}
  \caption{\textbf{Joint DD beats adapting the large model alone, for little extra compute.} Tent on ImageNet-C (15 corruptions, severity 5), episodic (reset at every shift, mean of 3 seeds): top-1 accuracy (\%) and GFLOPs of one adaptation step per image (batch 64). 
  The large model is ViT-B/16 and the small model is ResNet-50 or ConvNeXt-T. 
  Against Large Tent, our method costs 24--26\% more GFLOPs and gains 15.2 points with ResNet-50 (Joint DD) and 2.9 points with ConvNeXt-T (Indep. DD). 
  Best accuracy per Duo in bold.
  }
  \label{tab:compute_cost}
\end{table}

\label{app:compute}
\Cref{fig:acc_vs_flops_vitb_rn50,fig:acc_vs_flops_vitb_convnext} plot episodic accuracy against the compute of one adaptation step per image: the forward pass of every member, the backward pass of every adapting member, and the gate, counted with PyTorch's FLOP counter (one multiply-add counts as two FLOPs). 
A frozen sidekick adds about $25\%$ to the cost of the frozen ViT-B/16 (about $8$ and $9$ GFLOPs for the ResNet-50 and the ConvNeXt-T, against $34$ for the ViT-B/16), and adapting a member roughly doubles its cost, since Tent backpropagates through almost the whole network even though it updates only the normalization layers. 
The gate itself is negligible.

\Cref{tab:compute_cost} lists the accuracy and GFLOPs behind both figures.
On the ResNet-50 Duo, the frozen Duo is both cheaper and more accurate than the ViT-B/16 adapted with Tent ($46.1\%$ vs.\ $42.9\%$, with about $40\%$ fewer FLOPs), and adapting only the small member inside the Duo ($50.5\%$) beats every single model at lower cost than adapting the ViT-B/16 alone. 
Joint DD costs about $25\%$ more than single-model Tent on the ViT-B/16. 
On the ConvNeXt-T Duo, whose small member collapses, Independent DD is the most accurate configuration, $2.9$ points above the ViT-B/16 adapted alone for about $25\%$ more compute, whereas modes that adapt a single member inside the Duo fall below single-model Tent. 
These counts add up the work of both members. Since the members are independent until their logits are combined, they can run in parallel, on two devices or two streams, which we do not measure here.

\section{Frozen Duos: Per-corruption Results}
\label{app:frozen}
\Cref{tab:temperatures:frozen_bn__vitb16_rn50v1,tab:temperatures:frozen_bn__vitb16_convnextt} give the temperatures and per-corruption accuracy of both frozen Duos, summarized in \cref{tab:temperatures:duos}.

\paragraph{Combining within a corruption.}
\label{app:within}
On frozen Duos, we refit the two temperatures on every batch of $B$ consecutive samples, with labels and in-sample.
On the ResNet-50 Duo, this beats the corruption-wise oracle by $0.6$ points at $B = 16$ but by only $0.1$ at $B = 128$.
The ConvNeXt-T Duo shows the same pattern, and so does clean data, which has no shift to track, with an even larger gain at $B = 16$ ($0.9$ points on both Duos).
Since each batch is fitted and evaluated on the same samples, the small-batch gain comes from overfitting to each sample. 
Held-out TS is thus within about a point of the best batch-level combination ($0.8$ and $1.3$ points at $B = 128$).
These oracles fit the temperatures by negative log-likelihood, so they are reference points for combinations that use one temperature pair per corruption or per batch, such as our gate, rather than strict bounds on accuracy.

\section{Adapting Duos: Per-corruption Results}
\label{app:duos}
\Cref{tab:adaptation_modes:tent__vitb16_rn50v1_bn_gate_nnn_bt256_beta0p25-0p5_recpair,tab:adaptation_modes:tent__vitb16_convnextt_bn__worst_shift_gate_nnn_bt256_beta0p25-0p5_recpair} give the per-corruption results of \cref{sec:exp:duotta} for both Duos.
\paragraph{Effect of the gate per corruption.}
Independent DD is never more than $0.2$ points below Fixed TS on any corruption of either Duo.
In Joint DD the gate's weights also enter adaptation, and the effect can go either way: on the ResNet-50 Duo, Joint DD is $1.0$ point below Fixed TS on average and $4.7$ points below on Gaussian noise, while on the ConvNeXt-T Duo it is $5.5$ points above a Fixed TS Duo that itself falls below the frozen Duo ($38.6\%$ vs.\ $41.0\%$).
On the ConvNeXt-T Duo, Independent DD is the most accurate configuration ($45.8\%$, $2.6$ points above its Fixed TS counterpart).

\paragraph{The Duo entropy on the ConvNeXt-T Duo.}
When only ConvNeXt-T adapts on the Duo entropy, with the ViT-B/16 frozen, the Duo fails on the same corruptions and to nearly the same accuracy as when both members adapt: $2.4\%$ vs.\ $2.4\%$ on contrast, $8.7\%$ vs.\ $8.5\%$ on zoom blur and $11.6\%$ vs.\ $11.5\%$ on snow.
With both adapting, the ViT-B/16's learning rate is also small ($6.2\times10^{-6}$, \cref{tab:hparams_episodic}).
The shared objective thus harms this Duo mainly through the combination, which follows the collapsed sidekick, rather than by degrading the ViT-B/16.
A collapsed member is confident, so its scaled logits can dominate a fixed combination: the collapsed ConvNeXt-T overrides the ViT on $\sim 48\%$ of their disagreements, against $26\%$ for the collapsed ViT on the ResNet-50 Duo.

\tablepair[0.85]{tables/1_temperatures/tab_temperatures_frozen_bn_vitb16_rn50v1}{tables/1_temperatures/tab_temperatures__frozen_bn__vitb16_convnextt}

\tablepair{tables/2_adapting_duos/tab_adaptation_modes__tent__vitb16_rn50v1_bn_gate_nnn_bt256_beta0p25-0p5}{tables/2_adapting_duos/tab_adaptation_modes__tent__vitb16_convnextt_bn__worst_shift_gate_nnn_bt256_beta0p25-0p5}

\begin{table*}[t]
  \centering\footnotesize
  \setlength{\tabcolsep}{5pt}
  \caption{\textbf{Continual hyperparameters.} Learning rates of the large and small member, and our reset rule's per-member settings $(\alpha_{\mathrm f}, \alpha_{\mathrm s}, W_0, \delta)$.}
  \label{tab:hparams_continual}
  \begin{tabular}{@{}llcc cc@{}}
    \toprule
    & & \multicolumn{2}{c}{Learning rate} & \multicolumn{2}{c}{Reset rule $(\alpha_{\mathrm f}, \alpha_{\mathrm s}, W_0, \delta)$ (\cref{sub:method:reset})} \\
    \cmidrule(lr){3-4}\cmidrule(l){5-6}
    Setup & Reset & Large & Small & Large & Small \\
    \midrule
    \multicolumn{6}{@{}l}{\textit{ViT-B/16 + ResNet-50}} \\
    Tent, Independent & None         & 1.75e-5 & 3.1e-5  & -- & -- \\
                      & RDumb        & 7e-5    & 5e-4    & -- & -- \\
                      & SAR recovery & 5.6e-4  & 8.84e-5 & -- & -- \\
                      & ASR          & 5.6e-4  & 8.84e-5 & -- & -- \\
                      & DD (ours)    & 5.6e-4  & 8.84e-5 & $(0.05, 0.005, 20, 0.0875)$ & $(0.02, 0.005, 20, 0.175)$ \\[2pt]
    Tent, Joint       & None         & 2e-4    & 1.41e-3 & -- & -- \\
                      & RDumb        & 8e-4    & 4e-3    & -- & -- \\
                      & DD (ours)         & 1.6e-3  & 7.07e-4 & $(0.1, 0.01, 20, 1.05)$ & $(0.1, 0.025, 20, 1.4)$ \\[2pt]
    ETA, Independent  & None         & 6.25e-5 & 8.85e-5 & -- & -- \\
                      & RDumb        & 5e-4    & 2.5e-4  & -- & -- \\
                      & ASR          & 1e-3    & 2.5e-4  & -- & -- \\
                      & DD (ours)    & 5e-4    & 2.5e-4  & $(0.1, 0.025, 20, 0.05)$ & $(0.1, 0.025, 20, 0.05)$ \\[2pt]
    SAR, Independent  & RDumb        & 1e-3    & 1e-3    & -- & -- \\
                      & SAR recovery & 1e-3    & 2.5e-4  & -- & -- \\
                      & DD (ours)    & 1e-3    & 1e-3    & $(0.05, 0.005, 20, 0.35)$ & $(0.02, 0.005, 100, 0.7)$ \\
    \midrule
    \multicolumn{6}{@{}l}{\textit{ViT-B/16 + ConvNeXt-T}} \\
    Tent, Independent & None         & 2.45e-5 & 1.55e-6 & -- & -- \\
                      & RDumb        & 9.9e-5  & 2.19e-6 & -- & -- \\
                      & DD (ours)    & 3.96e-4 & 8.77e-6 & $(0.05, 0.005, 50, 0.5)$ & $(0.1, 0.01, 50, 0.25)$ \\[2pt]
    Tent, Joint       & None         & 4.37e-6 & 7e-5    & -- & -- \\
                      & RDumb        & 4.37e-6 & 7e-5    & -- & -- \\
                      & DD (ours)    & 6.19e-6 & 7e-5    & $(0.05, 0.005, 50, 1.0)$ & $(0.05, 0.005, 50, 1.0)$ \\
    \bottomrule
  \end{tabular}
\end{table*}

\end{document}